\RequirePackage{fix-cm}
\documentclass[smallextended]{svjour3}       
\smartqed  
\usepackage{graphicx}
\usepackage{caption}
\usepackage{subcaption} 
\usepackage[english]{babel}
\usepackage[utf8]{inputenc}
\usepackage{algorithm}
\usepackage{algpseudocode}
\usepackage{wrapfig}
\usepackage{amsmath}
\DeclareMathOperator*{\argmax}{arg\,max}

\usepackage[normalem]{ulem}
\usepackage{hyperref}
\hypersetup{colorlinks=false,breaklinks=true}
\usepackage{booktabs}
\usepackage{tabularx}
\usepackage{breakcites}
\usepackage[table]{xcolor}
\usepackage{multirow}
\usepackage{cite}

 \usepackage{mathptmx}      
\usepackage{latexsym}

\usepackage[section]{placeins}

\usepackage[utf8]{inputenc}
\usepackage[T1]{fontenc}
\usepackage{ltablex}
\usepackage{caption, booktabs}
\newcolumntype{s}{>{\columncolor{lightgray}} l}
\newcolumntype{k}{>{\centering\arraybackslash}p{2cm}}

\def\BibTeX{{\rm B\kern-.05em{\sc i\kern-.025em b}\kern-.08emT\kern-.1667em\lower.7ex\hbox{E}\kern-.125emX}}
\usepackage{xcolor}
\usepackage{longtable}
\begin{document}

\title{Co-eye: A Multi-resolution Symbolic Representation to Time Series Diversified Ensemble Classification}

\titlerunning{Co-eye}        

\author{Zahraa S. Abdallah        \and
        Mohamed Medhat Gaber 
}


\institute{School of Computing and Digital Technology, Birmingham City University, England , UK \\
Zahraa S. Abdallah\at
\email{zahraa.abdallah@bcu.ac.uk}
 \and 
Mohamed Medhat Gaber\at
\email{Mohamed.Gaber@bcu.ac.uk} 
}

\date{Received: date / Accepted: date}

\maketitle

\begin{abstract}
Time series classification (TSC) is a challenging task that attracted many researchers in the last few years. One main challenge in TSC is the diversity of domains where time series data come from. Thus, there is no ``one model that fits all'' in TSC. Some algorithms are very accurate in classifying a specific type of time series when the whole series is considered, while some only target the existence/non-existence of specific patterns/shapelets. Yet other techniques focus on the frequency of occurrences of discriminating patterns/features. This paper presents a new classification technique that addresses the inherent diversity problem in TSC using a nature-inspired method. The technique is stimulated by how flies look at the world through ``compound eyes'' that are made up of thousands of lenses, called ommatidia. Each ommatidium is an eye with its own lens, and thousands of them together create a broad field of vision. The developed technique similarly uses different lenses and representations to look at the time series, and then combines them for broader visibility. These lenses have been created through hyper-parameterisation of symbolic  representations (Piecewise Aggregate and Fourier approximations). The algorithm builds a random forest for each lens, then performs soft dynamic voting for classifying new instances using the most confident eyes, i.e, forests. We evaluate the new technique, coined Co-eye, using the recently released extended version of UCR archive, containing more than 100 datasets across a wide range of domains. The results show the benefits of bringing together different perspectives reflecting on the accuracy and robustness of Co-eye in comparison to other state-of-the-art techniques. 
\end{abstract}

\section{Introduction}
\begin{figure}[!h]
\centering
\begin{subfigure}{.7\textwidth}
  \centering
  \includegraphics[width=\linewidth]{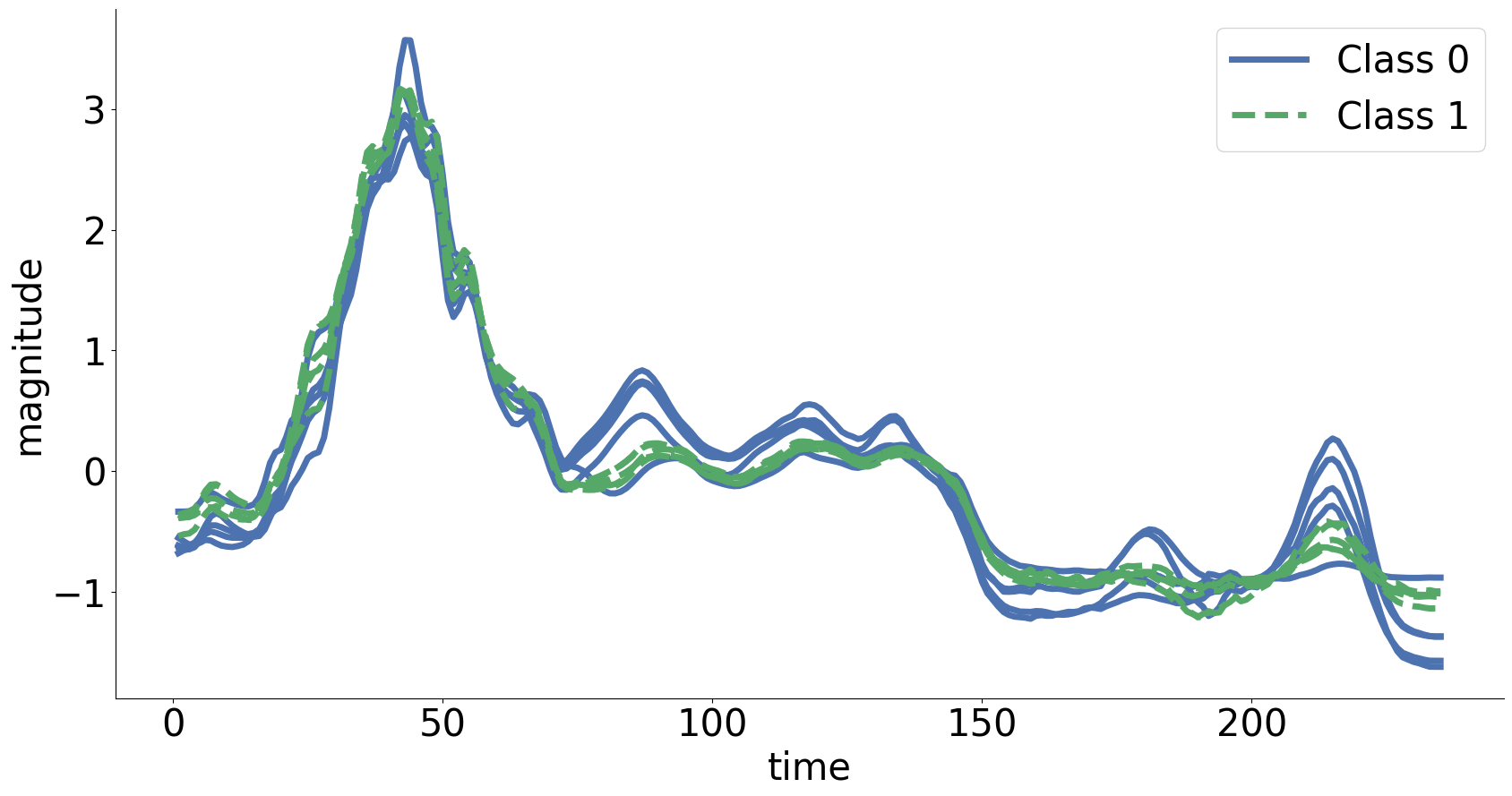}
  \caption{Spectrum time series (Strawberry)}
  \label{fig1:sub1}
 
\end{subfigure}
\begin{subfigure}{.7\textwidth}
  \centering
  \includegraphics[width= \linewidth]{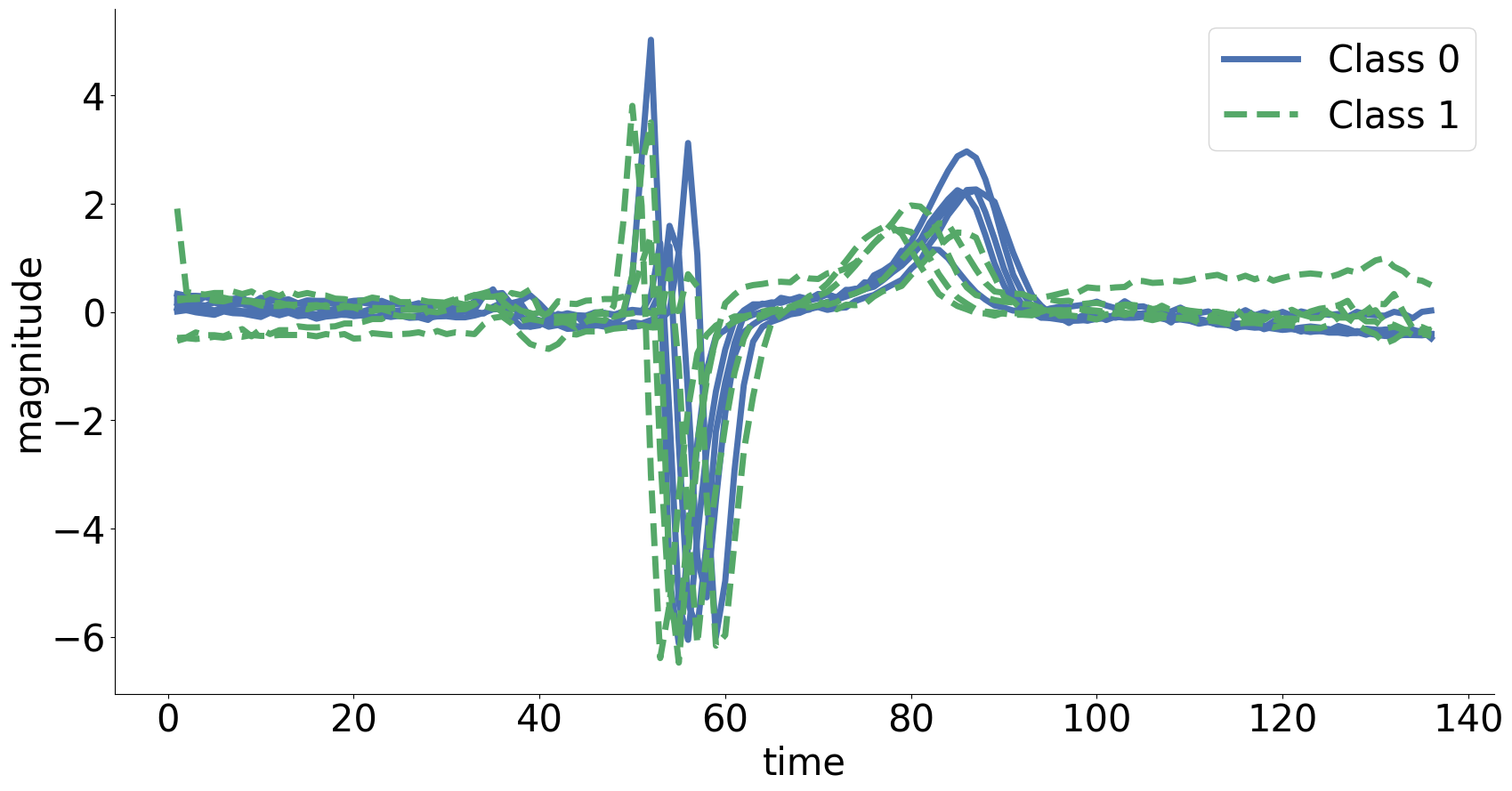}
  \caption{ECG time series (ECGFiveDays) }
  \label{fig1:sub2}
\end{subfigure}
\caption{\color{black}Samples of two different classes in spectrum and ECG time series demonstrating diversity in time series domains/shapes}
\label{ECGVSSpectrum}
\end{figure}

Time series classification (TSC) became a topic of great interest in the last few years. Accurate classification of time series can contribute to a variety of problems in a wide range of domains such as signal processing, pattern recognition, spectrum analysis, energy consumption analysis and many others. Notable algorithms have been developed to address the classification problem, while the vast majority of research has focused on developing similarity measures for accurate classification. A significant challenge that faces time series classification is the diversity of data that reflects the diversity of domains from-where data has been collected. \textcolor{black}{Time series of an electrocardiogram (ECG) in the medical domain}, for example, is significantly different from spectrum data \cite{Strawberry_dataset} as shown in Figure \ref{ECGVSSpectrum}. \textcolor{black}{Food spectrographs are used in chemometrics to classify food types, a task that has obvious applications in food safety and quality assurance. The classes in this dataset are strawberry (authentic samples) and non-strawberry (adulterated strawberries and other fruits). Obtained using Fourier transform infrared (FTIR) spectroscopy with attenuated total reflectance (ATR)  sampling.} \textcolor{black}{Both datasets, among others reported in this paper, are presented in \cite{UCR} and discussed in \cite{TSCwebsite,bagnall2017great}}.

\begin{figure}[ht!]
    \centering
    
\centering
    \includegraphics[width=0.7\textwidth]{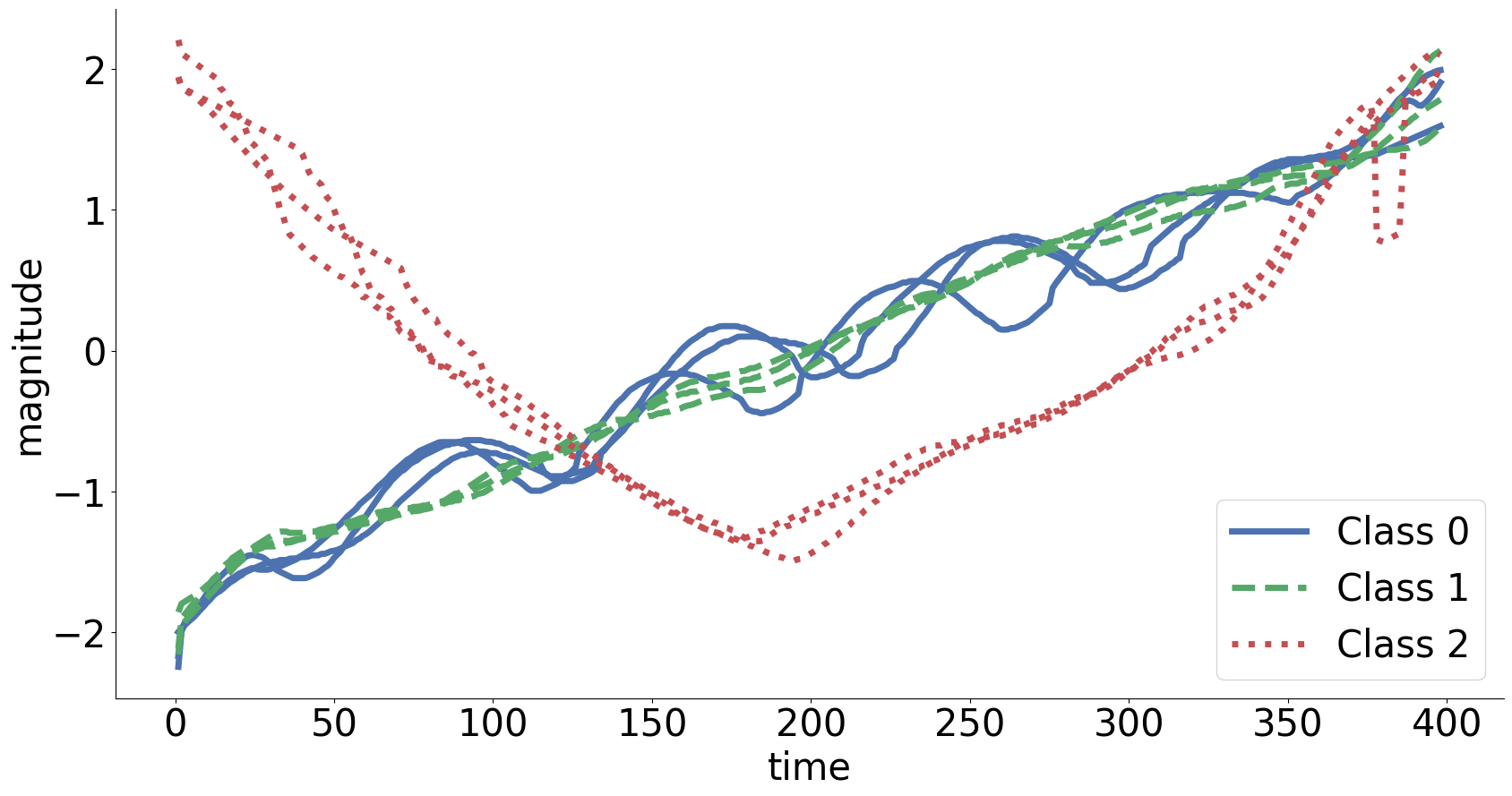}
    \caption{\textcolor{black}{Samples} of Symbol time series: an example of a whole series view }\label{symbols}
\end{figure}

TSC algorithms can be categorised based on the type of discriminatory features adopted for classification. Bagnall et al \cite{bagnall2017great} classified techniques as: whole series, intervals, shaplets, dictionary and combinations. Whole series techniques look at time series as a whole. The main focus of these techniques is to best align between series in order to find similarities. These techniques perform well with time series that has distinguished features concerning the entire series. A good example of that is Symbols dataset \textcolor{black}{\cite{UCR}. Figure \ref{symbols} illustrates the importance of a global view of this kind of series in order to find discriminatory features. The dataset is generated by asking thirteen people to copy the randomly appearing symbol as best they could. There were 3 possible symbols, each person contributed about 30 attempts. The data is the x-axis motion in drawing the shape. Figure \ref{symbols} represents the three possible symbols. As shown in the figure, a global view of the motion of drawing is more crucial for distinguishing between symbols than specific intervals observation.}

\begin{figure}[!ht]
\centering
    \includegraphics[width=0.7\textwidth]{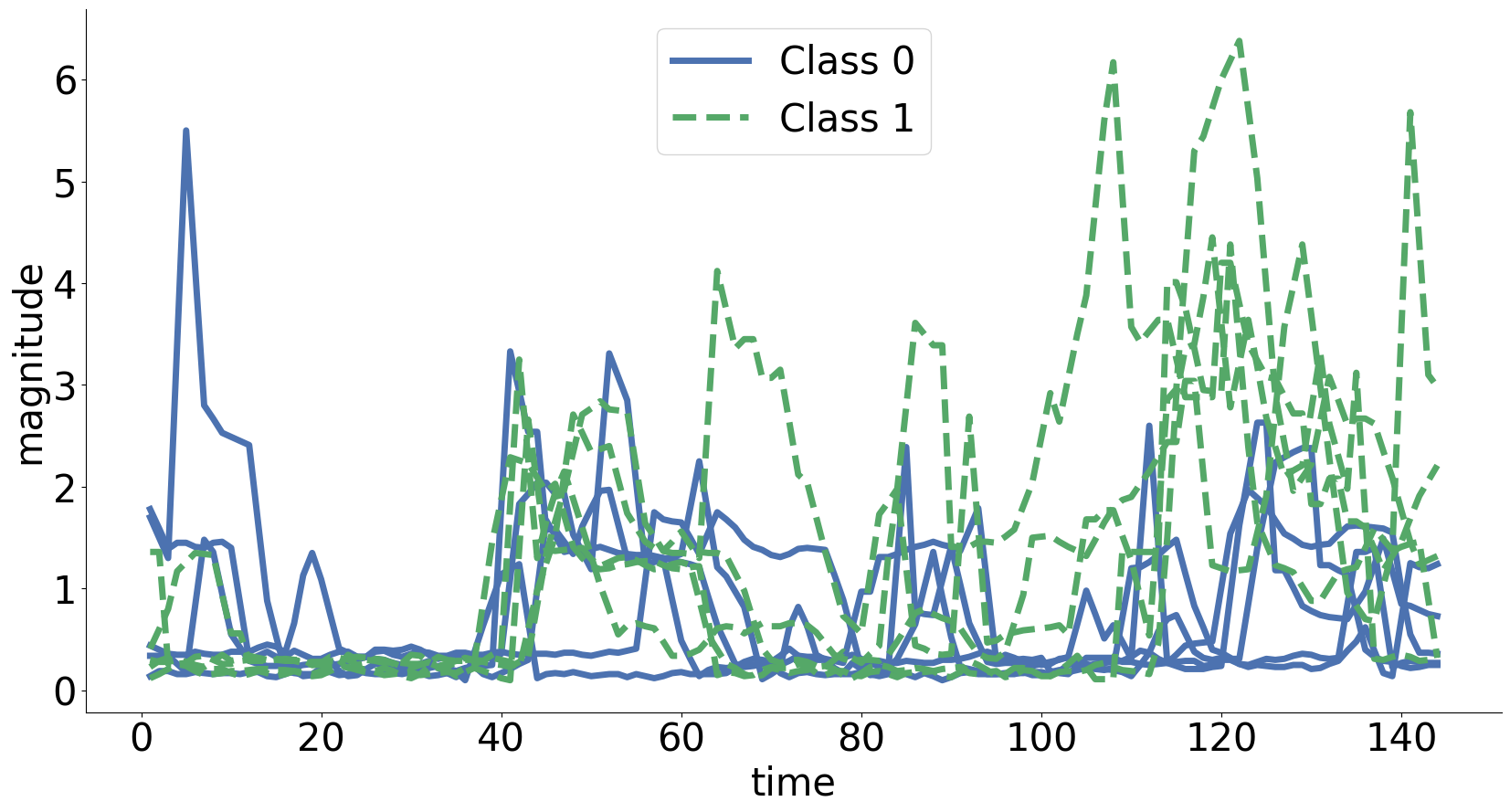}
    \caption{\textcolor{black}{Samples} of Powercons dataset: an example of intervals in time series}
    \label{powercons}
\end{figure}

Instead of examining the whole series, interval techniques select one or more phase dependent intervals of the series and extract features based on each. \textcolor{black}{The PowerCons dataset contains an individual household electric power consumption in one year distributed in two season classes: warm (class 1) and cold (class 2). The sampling rate is every ten-minute over a period of one year. As shown in Figure \ref{powercons}, the electric power consumption profiles differ markedly within classes. The PowerCons dataset is an example of data where interval techniques are expected to dominate as they can effectively capture different signatures of power consumption of different seasons. }

Shapelets are short phase dependent patterns that identify classes. Algorithms that rely on shapelets for classification look at the existence or absence of specific shapelets, while the shapelet actual location is irrelevant \textcolor{black}{\cite{grabocka2014learning}}. This could be very useful in finding abrupt change in signals such as ECG. Dictionary-based methods, on the other hand, capture the frequency of subsets of the series rather than their existence. \textcolor{black}{An example of Dictionary-based method is \textcolor{black}{ Bag-of-SFA-Symbols (BOSS)} \cite{schafer2015boss} that relies on a bag of words to build a TS dictionary}. From the aforementioned examples, it is clear that ``there is no one model that can fit all'' in TSC. Each data type has its own characteristics that give the superiority to one or more of these methods for an accurate classification. Thus, ensemble has become a very popular approach for improving the general accuracy. Some ensemble methods are based on same core classifiers such as Time Series Forest (TSF) \cite{deng2013time} and BOSS \cite{schafer2015boss}. While others fuse various stand-alone components of classifiers such as Collective of Transformation Ensembles (COTE) \cite{bagnall2015time} and \textcolor{black}{ensembles of elastic distance measures (EE) \cite{lines2015time}}.

\begin{wrapfigure}{R}{0.3\textwidth}
    \includegraphics[width=0.3\textwidth]{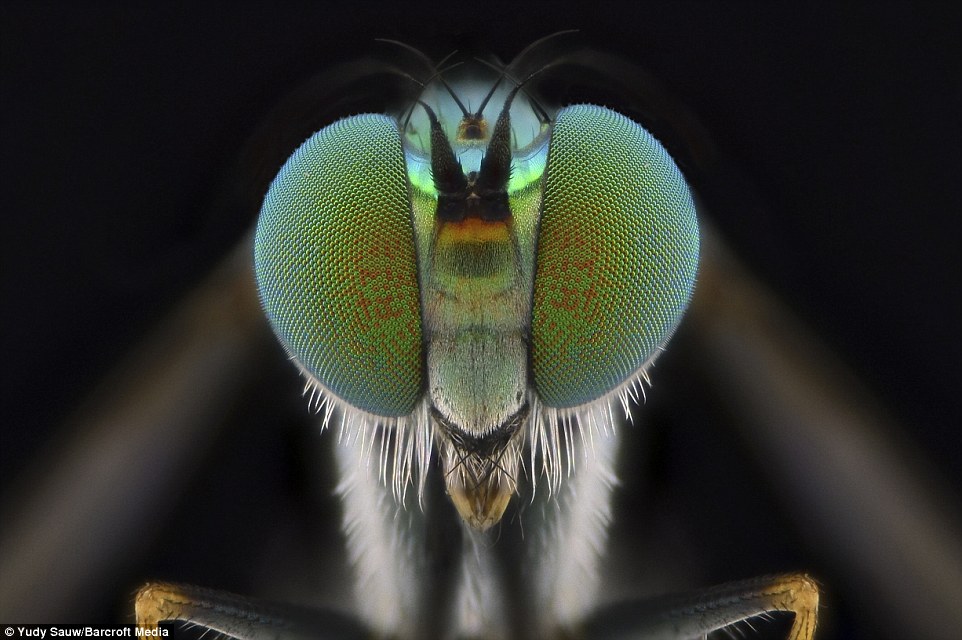}
     \caption{\textcolor{black}{A view of flies' compound eye (Source: Wikimedia Commons)}}
     \label{compeye}
\end{wrapfigure}

\begin{figure}[h!]
    \centering
    \includegraphics[width=0.7\textwidth]{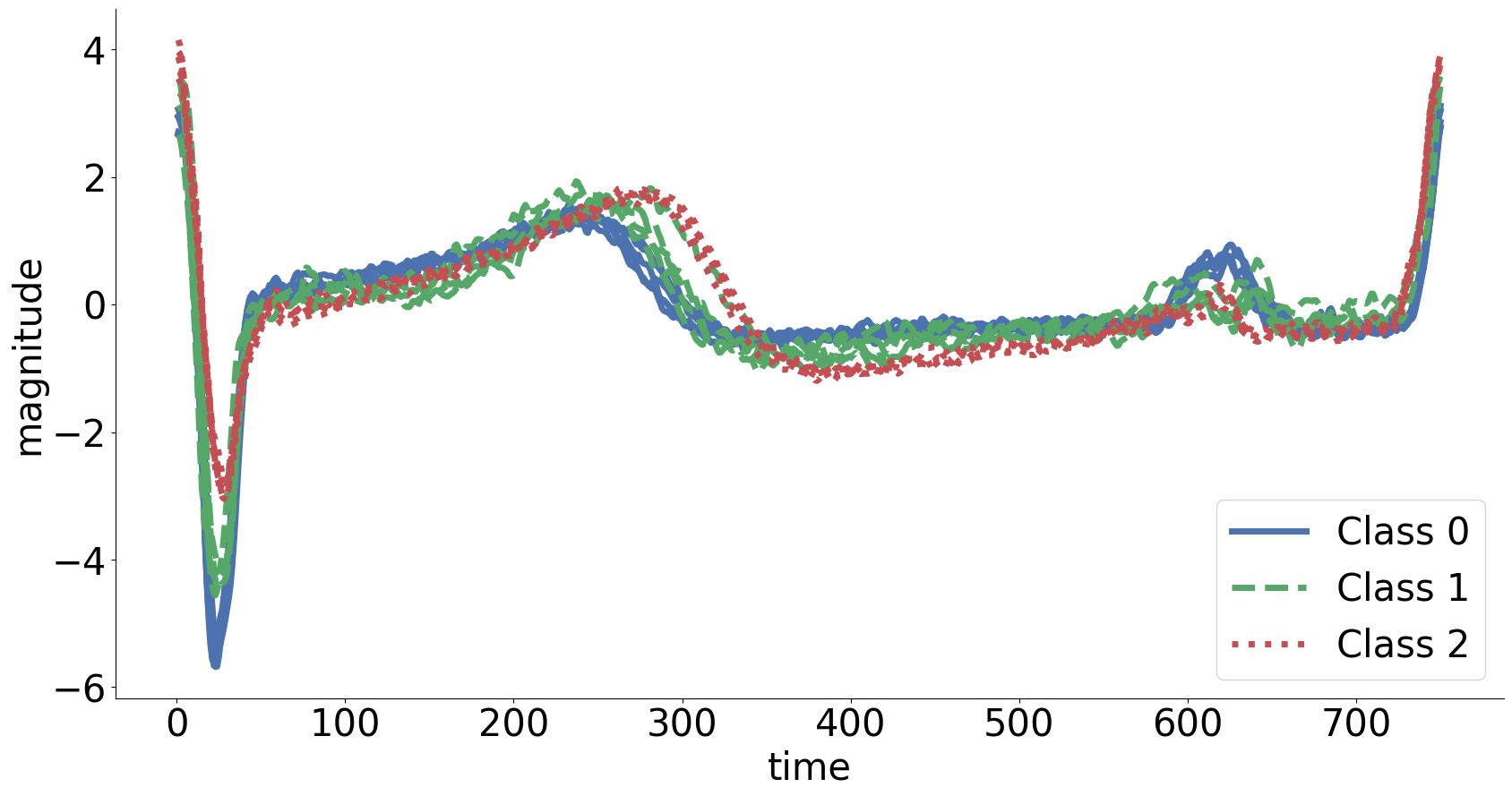}
    \caption{\color{black}A view of 10 samples from two classes in ``NonInvasiveFetalECGThorax1'' dataset }
    \label{FetalECG}
\end{figure}

In this paper, we propose a method that looks at time series from different perspectives similar to flies' compound eye. A combined eye consists of many ommatidia, each one is an individual eye by itself as shown in Figure \ref{compeye} (credited to Yudy Sauw). 
The technique combines time and frequency domains with various lenses in order to have a broader view of time series. Thus, the classification model is a collection of random forests, while each forest uses an individual lens. \textcolor{black}{Figure \ref{FetalECG} shows ten samples of three classes in ``NonInvasiveFetalECGThorax1'' UCR dataset \cite{UCR, silva2013noninvasive}. Each of this time series corresponds to the record of the ECG from the left and the right thorax. This series requires both a wider lens, for a global view, in addition to a fine-grained one in order to find distinguishing features in each class, note the fine change between the two classes (in the y-axis) around 60, 350 and 650 in time (x-axis). Relying only on one lens and ignoring others is likely to lead to inaccurate classification. Also, deciding how wide or narrow are the lenses is an important parameter in order to correctly capture the change in the series. Therefore, Co-eye has an advantage of combining various lenses together using hyper-parameterisation in order to decide the best lenses for accurate classification based on cross-validation of training data.} 

This technique is different from other previous work in many ways. First, it fuses both time and frequency symbolic representations of time series. Second, it represents a new dynamics of zooming in and out to establish a consolidated view of series by combining different granularities through hyper-parameterisation of each representation, i.e., lenses. Third, the lenses further diversify among the trees of the forest, through enrichment of the features. This gives the method an edge over other ensemble-based methods that operate over a fixed set of engineered features. Finally, the algorithm applies a dynamic voting mechanism to classify each individual time series based on the most confident forests/lenses amongst the collection of forests. Therefore, two different series \textcolor{black}{that belong} to the same class can be classified using two different sets of forests/lenses depending on the discriminating features in each series. 

This paper is organised as follows. Section \ref{relatedWork} discuses the related literature, Section \ref{background} provides a background for the proposed algorithm, while Section \ref{methodology} discusses Co-eye in details. In Section \ref{exp}, we evaluate Co-eye performance and analyse the results. The paper is concluded in Section \ref{con}.

\section{Related work}
\label{relatedWork}
Bagnall et al \cite{bagnall2015time} built the most complex ensemble in TSC, based on two observations: (1) improvements in TSC through transformations; and (2) the notable success in ensemble-based TSC methods, when using a particular transformation. \textcolor{black}{The COTE method showed superior performance over the other TSC methods at the cost of high complexity}. The proposed Co-eye method, on the other hand, makes use of these two observations, using a single \textcolor{black}{type of} classifier with different transformations resulting in a notably simpler ensemble than COTE, with an effective classification accuracy. \textcolor{black}{The high complexity of COTE stems from the fact that multiple types of classifiers are adopted including $k$-Nearest Neighbours ($k-NN$), Naive Bayes, decision tree, support vector machines with linear and quadratic basis function kernels, Random Forest (with 100 trees), Rotation Forest (with 10 trees) and a Bayesian network. The weighted voting is used to combine the results. COTE also used transformations in different domains. Hierarchical Vote
Collective of Transformation-based Ensembles (HIVE-COTE) is an extension of COTE, adding more features that have significantly improved its accuracy, but at the cost of an even more complex ensemble \cite{lines2016hive}.} 

Time series forest (TSF) has been proposed in \cite{deng2013time}, using a tree-based ensemble. Adopting interval-based features, and inspired by random forests, a randomisation of the extracted features from the intervals has been applied resulting in a linear feature space in the length of the series used in constructing each tree. A new splitting criterion at each node was used. Despite its success, it lacks the multi-resolution power, brought by the lenses, in the proposed Co-eye.   

Bag-of-SFA-Symbols (BOSS) has been proposed in \cite{schafer2015boss}. It uses 1-NN classification over transformed time series, adopting Symbolic
Fourier Approximation (SFA). A number of computational methods to speed up the transformation phase from loglinear in the window size to linear have been applied. Additionally, a noise elimination method was used. Also the adoption of a number of window sizes was used to apply an ensemble of 1-NN classifiers, one for each window size. Unlike BOSS that varies the window size, the proposed Co-eye varies the alphabet size and the word length to increase the diversity, and to induce multiple resolutions of the series. Additionally, both \textcolor{black}{Symbolic Approximation Transformation (SAX) \cite{senin2013sax}} and SFA were used, increasing the number of lenses/features used to build the ensemble of trees, having multiple trees for each pair of word lengths and alphabet sizes. 

Lines and Bagnall \cite{lines2015time} have carried out an extensive experimental work to test state-of-the-art distance measures in TSC. The experiments showed that ensembling over a variety of distance measures consistently result in an accuracy boost in TSC. Although the proposed ensembles are quite different than the ensemble and fusion methods proposed in Co-eye, the work evidences the need to have a multi-resolution representation of the time series. In Co-eye, this was achieved through variations in the hyper-parameterisation of symbolic approximations, instead of applying a variety of distance measures as in Lines and Bagnall's work. 

Evidencing the need for varying the granularity of representation in TSC, when applied to long sequences, Lin et al \cite{lin2012rotation} showed a text-based inspired feature extraction method, namely, bag-of-words, resulting in a rotation invariant representation of the time series. Another bag of features representation was proposed in \cite{baydogan2013bag}. Drawing from random locations in the time series, multiple subsequences are extracted and shorter intervals are used as features. The features are then labelled, and summary statistics for subsequence labels for each feature is maintained. Other global features are added to train a classifier (Random Forest and SVM). The new representation shows efficacy, however, and unlike Co-eye, the multi-resolution is not fully explored. Similarly, Senin and Malinchik \cite{senin2013sax} have used Bag of SAX words, extracted using a sliding widow over each time series, in a vector space of class-dependent modelling of the corpus of SAX words. The inference is done through the application of cosine distance measure between an unlabelled time series, and each class represented in a vector space of SAX words. The so called SAX-VSM shows that a pattern-based representation of the series can give the method an edge over distance-based methods in a variety of domains. Despite being different, Co-eye also uses a pattern-based representation through variations in the hyper-parameterisation of both SAX and SFA representations.  

\textcolor{black}{More recently, deep learning methods have been adopted in TSC. A number of deep and shallow neural network architectures have been experimented with and compared to state-of-the-art \cite{fawaz2019deep}. The experiments showed the merit of residual neural networks, when compared with other architectures such as convolutional neural networks, and multi-layer perceptron. However, when best performing DNNs compared with other state-of-the-art methods, COTE and HIVE-COTE have shown superior performance \cite{fawaz2019deep}. As such, our discussion in this paper will be focused on non deep neural network methods.}

Having discussed related work in this section, it appears that various work on time series representation has contributed to boosting up the accuracy in TSC. Also it is evident that using ensembles instead of single classifiers has shown a superiority in accuracy. However, none of the previous work explored a systemic use of diversification of time series representation to boost up the classification accuracy of ensembles in TSC. Additionally, the hyper-parameterisation of the two symbolic representation (SAX and SFA) to generate a multi-resolution time series representation has not been exploited. Thus, the proposed Co-eye brings together these missing features, in a quest to further boost up the accuracy of TSC.    

\section {Background}
\label{background}
Before we get into Co-eye details, we present techniques that Co-eye utilises. One main block in Figure \ref{overview}, which outlines the Co-eye's overall process, is ``transformation''. This block transforms time series to a multi-resolution symbolic representation in order to create the diverse lenses of the compound eye. Various techniques in TSC leverage symbolic representation, because it provides a significant dimensional reduction, which enables a wider range of similarity measures to be applied. Also, transforming time series to a shorter string of symbols enables techniques from other domains, such as text mining and bioinformatics, to be applied effectively to time series classification. Two symbolic representations are well-studied in the literature and proven to be effective: (1) SAX, Symbolic Aggregate approXimation \cite{patel2002mining} and  (2) SFA, Symbolic Fourier Approximation \cite{schafer2012sfa}. 

\paragraph{Symbolic Aggregate Approximation (SAX)}
Consider a time series $TS$ that is a sequence of $n$ time dependent values. $TS$= ($t_1$, $t_2$, $\dots$, $t_n$). SAX transforms $TS$ to a string of length $w$, where $w<<n$. SAX transformation consists of two steps. First, the time series is normalised using $z$-normalisation, with a mean of 0 and a standard deviation of 1. The normalised $TS$ is transformed to SAX by applying Piecewise Aggregate Approximation (PAA) \cite{keogh2000simple}. Two parameters are required for PAA, word length $w$ and alphabet size $\alpha$. PAA divides the normalised time series into $w$ equally sized segments, then the mean value of each segment is computed. The sequence of $w$ mean values is transformed to a string of alphabet size $\alpha$ using a look-up table. \textcolor{black}{It is worth noting that Discrete Haar Wavelet Transform (DWT) can be identical to PAA when the time series length is an integral power of two. However, PAA is much faster to compute, and can handle time series of arbitrary length \cite{keogh2001dimensionality}.}

\begin{figure}[!h]
    \centering
    \includegraphics[width=0.7\textwidth]{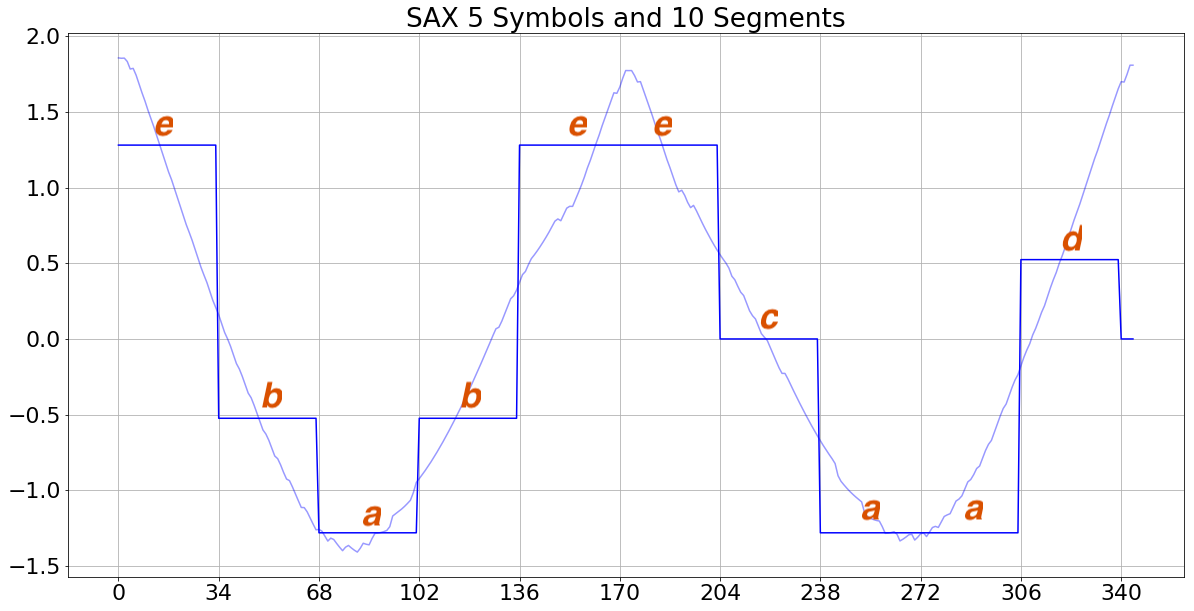}
    \caption{An illustration example of SAX transformation}
    \label{SAX}
\end{figure}

SAX creates its look-up table by creating equal-sized areas that are slicing the under-the-Gaussian-curve area. The $x$ coordinates of \textcolor{black}{these lines are called cuts}. By assigning a corresponding alphabet symbol to each interval between cuts, SAX performs the conversion of the PAA vector of segments to a string. Figure \ref{SAX} shows an illustrative example on ``DiatomSizeReduction'' UCR dataset transformed to PAA and then SAX of word length 10 and \textcolor{black}{alphabet with size 5}. The output string of this series is ``ebabeecaad''.

\paragraph{Symbolic Fourier Approximation (SFA)}\color{black}{is another symbolic representation of time series that is applied in the frequency domain, in contrast with SAX which is time-dependent. SFA approximation has two consecutive steps: approximation and quantisation. First, the normalised time series is approximated using low pass filtering, i.e., discrete Fourier transform (DFT). Word length $w$ is an important parameter in this step as it specifies the bandwidth of DFT, and consequently the number of Fourier coefficients produced in the approximation. Then, Fourier coefficients are transformed into a string representation using Multiple Coefficient Binning (MCB) \cite{schafer2015boss} in the quantisation step. MCB requires the alphabet size $\alpha$ which specifies the degree of quantisation for Fourier coefficients. SFA word is obtained using a look-up table of MCB intervals. 

SFA can be formalised  as follows. SFA aims to present each time series $TS$ as a string of symbols $s$ of length $w$. Hence, $SFA (TS)= s_1,s_2, \dots ,s_w$. In the approximation step, $TS$ of length $n$ is approximated where $DFT(TS)= f_1, f_2,\dots, f_w$, where each $f$  contains both real and imaginary values of Fourier transformation. \\

$DFT(TS)$ = $ \left( \begin{array}{ccc}
DFT (TS_1) \\
DFT(TS_2) \\
\dots  \\
DFT(TS_n) \\ \end{array} \right)$
= $ \left( \begin{array}{ccccc}
real_{11}  &img_{11}& ...& real_{1\dfrac{w}{2}}& img_{1\dfrac{w}{2}} \\
\dots &\dots& \dots&\dots&\dots \\
real_{n1}  &img_{n1}& ..& real_{n\dfrac{w}{2}}& img_{n\dfrac{w}{2}} \\ \end{array} \right)$ 
= $(f_1, f2,\dots f_w )$
\newline

In the quantisation step, MCB maps Fourier values ($f_1, f_2,\dots, f_w$) to a string of symbols of length $w$ and alphabet size $\alpha$. MCB first determines breakpoints for each $f$ by applying binning with equal-depth. MCB then labels each bin/interval by assigning the corresponding symbol using a look-up table. The table of labelled intervals in MCB is computed based on the training data.}
\color{black}
\section{Co-eye}
\label{methodology}
This section discusses in details the new ensemble method, Co-eye. First, we define some essential terminologies that we use throughout the following sections. Then an overview of Co-eye is depicted, followed by a detailed explanation of each component. 

Throughout this paper, we use ``word size'' symbol $w$ to refer to word length in Symbolic Aggregation approximation (SAX), and number of \textcolor{black}{coefficients}, as a reflection of word length, in Symbolic Fourier representation (SFA). We first define what the lens is. 

\begin{definition}
\textbf{Lens} ($l$) is a triplet representing the parameterised symbolic approximation used $<s,\alpha,w>$, where $s = \begin{cases}
      0, & \text{if SAX is used} \\
      1, & \text{if SFA is used}
    \end{cases}$, $\alpha$ is the alphabet size, and $w$ is the word size (in SAX) / Number of Fourier Coefficients (in SFA).  
\end{definition}

Each symbolic representation generates multiple lenses ($l$). An eye ($c$) is built for each lens, while a collection of eyes forms a compound eye. 

\begin{definition}
\textbf{Eye} is a classifier $c_i$ trained using a lens $l_i$ that has specific values for the triplet $<s=s_i, \alpha=\alpha_i, w=w_i>$.
\end{definition}

\begin{definition}
\textbf{Co-eye} is an ensemble of classifiers $C=\{c_1,c_2,\dots,c_k\}$, where $k$ is the total number of classifiers in the ensemble, and $\forall c_i \in C$, a correspondent lens $l_i$ is used for training the classifier $c_i$, collectively forming a set of lenses $L=\{l_1,l_2,\dots,l_k\}$. 
\end{definition}

\begin{figure}[!h]
    \centering
    \includegraphics[width=1.1\textwidth]{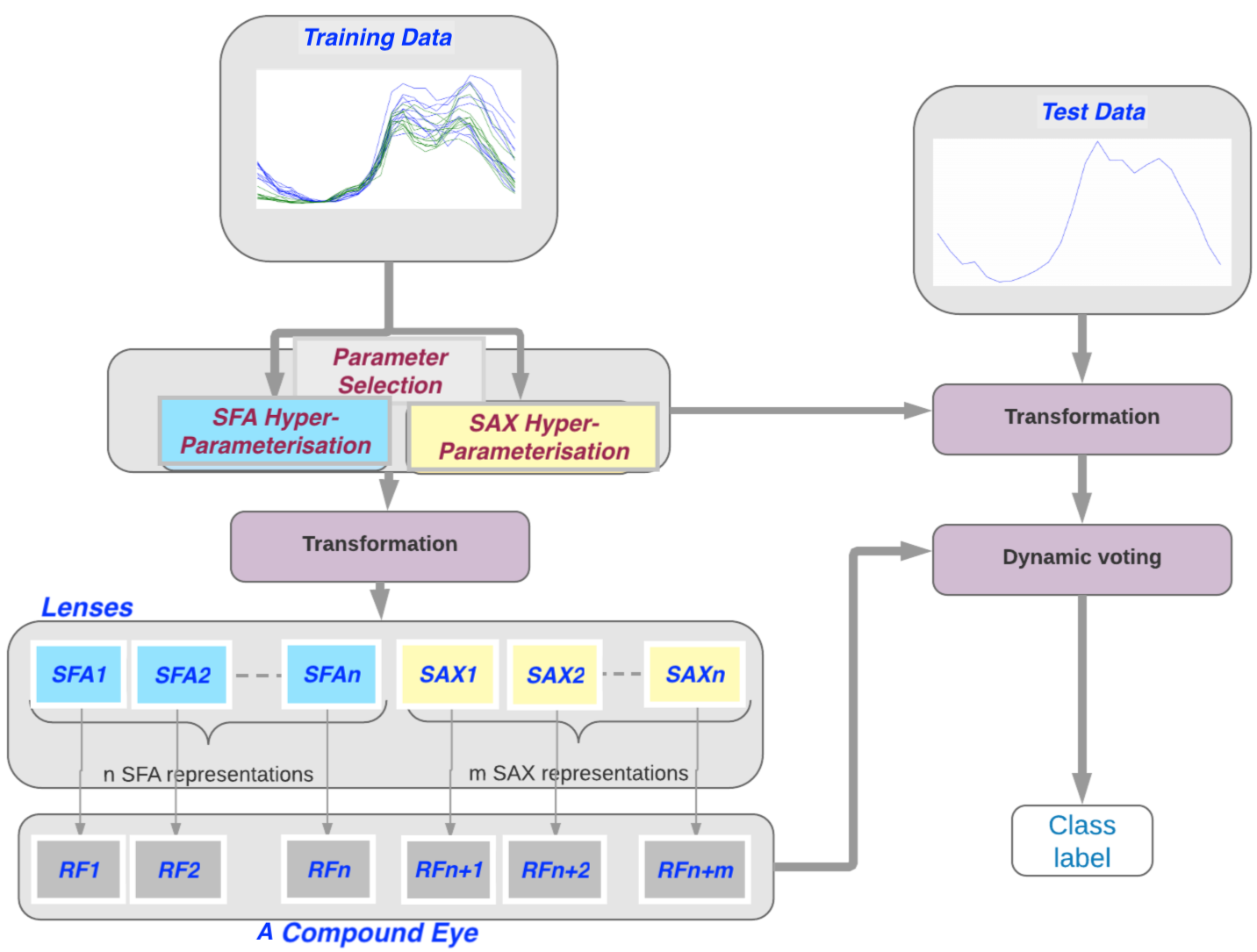}
    \caption{An overview of Co-eye Training and testing phases}
    \label{overview}
\end{figure}

Based on these definitions,  Figure \ref{overview} shows an outline of Co-eye algorithm. Co-eye consists of two phases; training and classification, as in most standard classification techniques. In the training phase, labelled data is transformed to many different representations using both SAX and SFA. These presentations (lenses) are selected with hyper-parameterisation for SAX and SFA separately in order to choose the best set of parameters for each. The main intuition behind the Co-eye is to select the best set of parameters $w$ and $\alpha$ while zooming in, with short segments and long alphabets, and zooming out, with long segments and short alphabets. The same concept applies for hyper-parameterisation in the frequency domain using SFA. We then build a Random Forest (Eye) for each transformed representation. The classification model in Co-eye is a collection of forests of symbolic representations in both SFA and SAX. \textcolor{black}{Unlike multi-view learning that relies on creating views of data, and has been recently adopted in TSC \cite{li2016multi}, Co-eye creates multi-resolution of time series. The main difference is that a data view can be any data representation that creates diversity such as subspacing. It stemmed from work in semi-supervised learning, instead of supervised learning based ensembles.}

Random Forest\cite{RandomForest} are typically diverse and do not overfit with the increase in the number of trees in the forest. These two features are coherent with Co-eye mechanism and objectives. The diversity of random forest is due to random samples selected for each tree using bootstrap sampling, and at each node using splitting over a random feature subspace (typically the size of the subspace is equal to \textcolor{black}{the} square root of the total number of features). Random forest mitigates the overfitting by adding more trees, which produces a limiting value of generalisation error. 

To classify unlabelled series, Co-eye first transforms series into the same set of representations. Then, soft and dynamic voting is performed to choose amongst the most confident forests. 

\subsection{Training phase}
Algorithm \ref{trainingAlg} depicts the outline of Co-eye training phase. Parameters in Co-eye are generated automatically using \textcolor{black}{$SearchLenses$} method which implements hyper parameterisation for both symbolic representations, SAX and SFA (lines 2 and 3). The output of the hyper-parameterisation step is a set of selected pairs/lenses of $w$, word length, and $\alpha$, alphabet size. Details of \textcolor{black}{$SearchLenses$} are discussed in the next section. Co-eye transforms the time series to a symbolic representation for each selected pair, then builds an eye using random forest on the transformed series (lines 5-6 for SAX pairs, 8-9 for SFA pairs). The final classification model contains $M$+$N$ random forests (line 12), where $M$ is the number of SAX pairs and $N$ is the number of SFA pairs. 

\begin{algorithm}[!h]
\caption{Training Phase}\label{trainingAlg}
\begin{algorithmic}[1]
\Procedure{buildClassifierCo-eye} {$TS$}\Comment{TS is the training data of length $n$}

\State $Pairs_{SAX} \gets searchLenses_{SAX}(TS)$
\State $Pairs_{SFA} \gets searchLenses_{SFA}(TS)$

\For {\texttt{$\alpha$ and $w$ in $Pairs_{SAX}$}}
    \State \textit{$SAX_{\alpha, w} \gets$ symbAggAppx($TS$, $\alpha$, $w$)}
    \State \textit{$clfSAX_{\alpha , w} \gets $ RandomForest($SAX_{\alpha, w}$)}
\EndFor
\For {\texttt{$\alpha$ and $w$ in $Pairs_{SFA}$}}
    \State \textit{$sFA_{\alpha , w} \gets$ symbFourierAppx($TS$, $\alpha$, $w$)}
    \State \textit{$clfSFA_{\alpha , w} \gets $ RandomForest($SFA_{\alpha , w}$)}
\EndFor

\State $ClfModel \gets fuse(clfSAX, clfSFA)$
\EndProcedure
\end{algorithmic}
\end{algorithm}
\subsection{Co-eye hyper-parameterisation}
\label{hyper-param}
\begin{figure}[ht!]
    \centering
    \includegraphics[width= \textwidth]{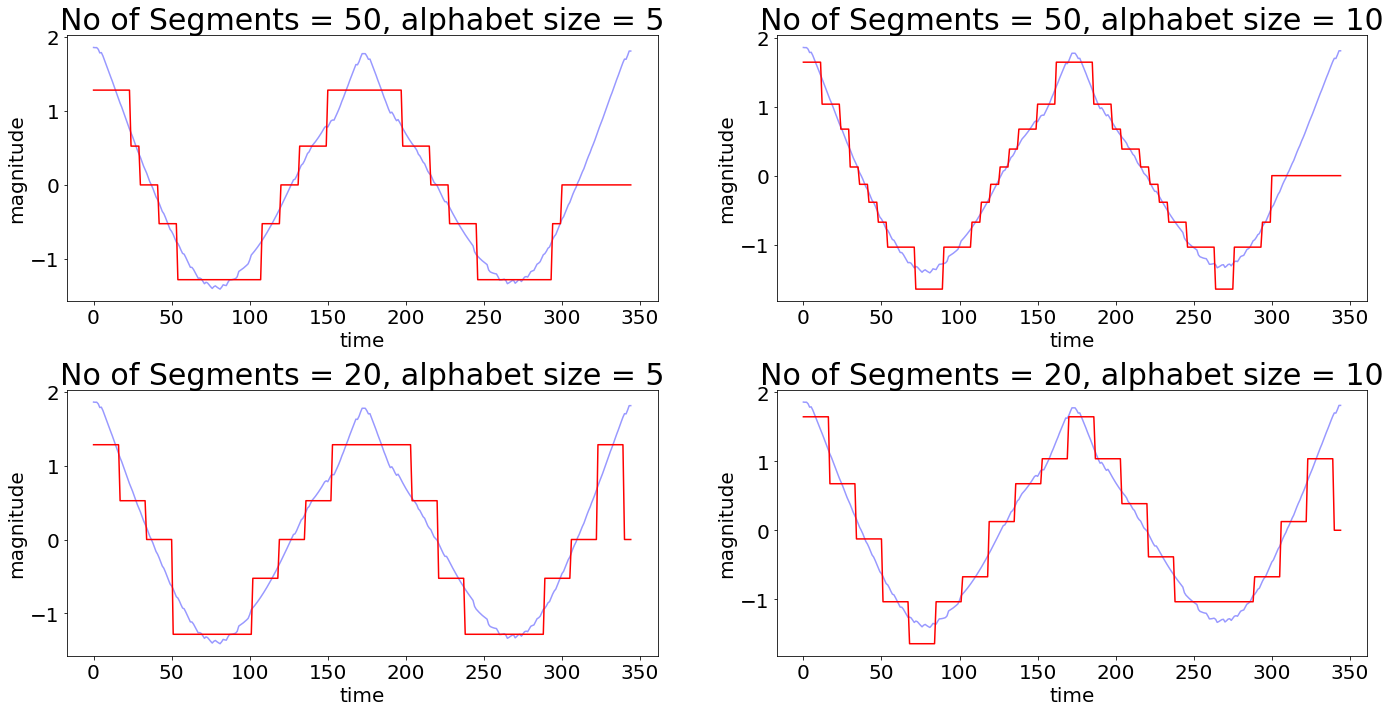}
    \caption{SAX parameters}
    \label{saxparam}
\end{figure}

As discussed in the background, symbolic representations require at least two parameters as an input, typically word length and alphabet size. To the best of our knowledge, there is no best selection for these parameters. Researchers tend to use optimisation methods, such as DIRECT \cite{finkel2003direct} to address the selection problem. However, one optimal selection may not offer the most efficient solution for an accurate representation. TSC typically requires multi-resolution representation with various combinations of parameters. Figure \ref{saxparam} shows an example of time series transformed with SAX using 4 different sets of parameters. Both word length, represented as the number of segments, and alphabet size determine the granularity of approximation. Very high-resolution lens, in this context, uses a longer \textcolor{black}{word length} $m$ and/or a \textcolor{black}{larger} alphabet size $\alpha$. This sharp lens is very important to spot small changes in the series revealing patterns of motifs and discords. However, a global view of the time series is as important too. Thus, a wider lens, represented by a shorter symbolic presentation and/or a small alphabet size, explores the global patterns in the series. The key feature of Co-eye is to combine various lenses in order to discover local and global discriminating features with multi-resolution symbolic representations.  

Co-eye selects the best parameters, which reflect best lenses, to look at time series based on cross-validation on the training data. This mechanism is applied on both time and frequency symbolic representations, i.e., SAX and SFA. Algorithm \ref{paramAlg} explains the process of finding the best lenses for Co-eye. The aim of this step is to find the best pairs of $w$ and $\alpha$ based on the training data. Specifying these lenses is the key to build an accurate and robust classification model. \textcolor{black}{In order to find the most accurate lenses, Co-eye starts with looking for regions of best pairs of word length $w$ and alphabet size $\alpha$. The upper bound of $\alpha$ is 26 (alphabet size), while the word length upper bound is the length of the time series with some margin. As the upper bound of the alphabet size is definite for all time series, we fix the alphabet selection first. Thus, for each alphabet, we aim \textcolor{black}{at finding} the best word length that offers the most accurate view of the series.} 

The training data is transformed into symbolic approximation using the two parameters $\alpha$ and $w$ (line 4). Then, accuracy on the training data is computed using cross validation (line 5).  The selection ends with choosing all word sizes that attain the maximum accuracy or very close to it (with 1\% margin) (line 8-9). \textcolor{black}{The margin allows for selecting all possible word lengths that attain  a good visibility region for a given alphabet. This process is repeated for all alphabets.}

\begin{algorithm}
\caption{SearchLenses algorithm}\label{param}
\label{paramAlg}
\begin{algorithmic}[1]
\Procedure{SearchLenses} {$TS$}\Comment{TS is the training data of length $n$}
\For {\texttt{$\alpha$ in $alphas$}}
    \For {\texttt{$w$ in $wordLengths$}}
        \State \textit{$symbTS_{\alpha , w} \gets$ symbAppx($TS$, $\alpha$, $w$)}
        \State \textit{$acc_{\alpha , w} \gets $ RandomForest($symbTS_{\alpha, w}$)}
        \State $accumlate(acc_{ \alpha , w}, acc_{all})$
    \EndFor
   \State $Threshold \gets max(acc_{all})- 0.01$
    \State $selPairs \gets filterPairs(Threshold, pairs)$ 
\EndFor

\EndProcedure
\end{algorithmic}
\end{algorithm}

The outcome of the hyper-parameterisation step \textcolor{black}{is composed of two sets} of lenses in both frequency and time domains. The collection of selected pairs are used to build the classification model (eyes) as explained in Algorithm \ref{trainingAlg}. 

\subsection{Classification phase}
The aim of this phase is to assign a class label to unlabelled series using the classification model built in the training phase. Algorithm \ref{classification} explains the classification process. The series is transformed to $N$ +$M$ symbolic representations (lines 3 and 7), each is classified using its corresponding forest (lines 4 and 8). The output of each forest is the prediction probability for all classes. The consolidated output is a probability matrix $predProb$ of $k$ $\times$ $c$, where $k$= $N$+$M$, $N$ is the number of SAX forests produced from $N$ SAX pairs, i.e., lenses and $M$ is the number of SFA forests produced from $M$ SFA lenses and $c$ is the number of classes. The label selection in Co-eye classifier votes amongst the most confident trees with the highest probability through soft voting, while the weight of each vote is the selection confidence (line 10).  

\begin{algorithm}
\caption{Classification Phase}\label{classification}
\begin{algorithmic}[1]
\Procedure{ClassifyInstance} {$T$, $clfModel$, $Pairs$}
\newline \Comment{T is unlabelled time series of length n}

\For {\texttt{$\alpha$ and $w$ in $Pairs_{SAX}$}}
    \State \textit{$SAX_{\alpha , w} \gets$ symbAggAppx($T$, $\alpha$, $w$)}
    \State \textit{$predProb \gets $ Classify ($SAX_{\alpha , w}$, $clfModel_{\alpha, w}$)}
\EndFor
\For {\texttt{$\alpha$ and $w$ in $Pairs_{SFA}$}}
    \State \textit{$SFA_{\alpha , w} \gets$ symbFourierAppx($T$, $\alpha$, $w$)}
    \State \textit{$predProb \gets $ Classify ($SFA_{\alpha , w}$, $clfModel_{\alpha , w}$)}
\EndFor

\State $classLabel \gets Vote(predProb)$
\EndProcedure
\end{algorithmic}
\end{algorithm}

The prediction matrix $predProp$ for $k$ random forests across $c$ class labels is as follows:  

$predProb$ = $ \left( \begin{array}{ccc}
P_{(1,1)}  & \dots& P_{(1,c)} \\
P_{(2,1)}  &\dots& P_{(2,c)} \\
\dots &\dots& \dots \\
P_{(k,1)}  &\dots& P_{(k,c)} \\ \end{array} \right)$
\newline
\hfill

Where $P_{(i,j)}$ is the prediction probability, i.e., confidence, of Random Forest $i$ for class label $j$. The matrix holds the prediction probability of eyes in order, i.e., $M$ followed by $N$. 

In order to choose the most confident label for a time series , we look through the most confident lenses/forests in $predProb$ matrix for each representation. Thus, we find the maximum confidence in each representation and its corresponding label $$ ConfLabel=  \forall_{i \in R} \argmax_{n=1}^{c} P_{(i,n)} , R\in [N,M] $$

$confLabels$ holds the most confident labels for each representation. If both representation agrees on a label, then it is assigned as the predicted label with confidence. In the case of disagreement, another round of voting is performed on second best confident labels between the two representations. Voting between only a representative of each transformation contributes in reducing bias that is possibly created due the number of pairs generated for each symbolic representation. For example, if SAX generated 50 pairs, while SFA generated only 10, the normal voting might be biased towards the larger pool of pairs. However, with this alternative mechanism, we choose only the most confident for each symbolic representation and vote between them. 

For an individual representation, many lenses might have the best confidence. If they all vote for one label then we choose any randomly. If there is a dispute,  the best label is the most frequent one, while the second best is the less frequent, having the same confidence. In case of a dispute with a tie (they both have the same frequency), then we choose any of them randomly.

Consider the following illustrative example of $confLabels$ for 5 random forests and two predicted class labels. The first 2 forests correspond to SAX lenses, the later 3 are for SFA.

$confLabels$ = $ \left( \begin{array}{ccccc}
SAX&SAX&SFA&SFA&SFA\\
C_1&C_2&C_1&C_2&C_1\\
0.8&0.9&0.8&0.6&0.7\\
 \\ \end{array} \right)$

The most confident label in the first two forests, that correspond to SAX, is $C_2$ with confidence 90\%. While the most confident in SFA is $C_1$ with 80\% confidence. As the two representations have no agreement, the second best confident label is considered for another round of vote. Both representations agree on $C_1$ in the second round, hence $C_1$ is selected in this case. 

This voting mechanism also gives flexibility for each time series to select the most confident forests/eyes in order to extract discriminating features for a specific series. Therefore, it enables dynamic matching of lenses/forests and series. \textcolor{black}{The proposed voting mechanism is instance driven, which is different than typical ensemble fuser mechanisms \cite{wozniak2014survey}. The mechanism proposed in this work belongs to the class label fusion category. Unanimous, simple majority and majority are among the common methods in this category. The weighted majority is adopted in this work, but on a selection of confident classifiers, instead of the whole ensemble.}

\section {Evaluation}
\label{exp}
\textcolor{black}{In this section, we discuss the set of experiments assessing Co-eye performance}. We first illustrate the experimental setup in \ref{expsetup} followed by analysis of parameter selections of Co-eye in \ref{parameters}. Details of Co-eye classification accuracy on UCR datasets are discussed in \ref{co-eye_acc}. Finally, we illustrate \textcolor{black}{the} Co-eye mechanism on a case study in \ref{casestudy}.  

\subsection{Experimental Setup}
\label{expsetup}
We evaluate Co-eye using the extended UCR Time Series Archive, published in 2018 \cite{UCR, TSCwebsite}. Since 2002, UCR archive has become an important resource in the time series research community. The new expansion of UCR increased the number of datasets from 85 to 128 by adding more realistic datasets with \textcolor{black}{larger} size and fewer labelled data. We use in these experiments the extended version excluding varied length time series (a total of 114 datasets). \textcolor{black}{The reported performance of all methods used in comparison with Co-eye throughout the paper is the published results in \cite{UCR, TSCwebsite} }

We train/test on the provided split data. The lenses selection is performed on the training data, and then assessing the accuracy is performed on the test data. 
We perform 5-fold cross-validation on the training data to define lenses for both SAX and SFA, i.e., a set of pairs of $w$, word length and $\alpha$, alphabet size. \textcolor{black}{Some datasets, such as Fungi, have only one example per class,} therefore we use leave-one-out cross-validation (LOO) instead of cross-validation. The number of estimators in Random Forest is set to 100 trees with ``Gini impurity'' function to measure the quality of a split. 

In Fourier transform, coefficients dictate the word length. The selection of Fourier Coefficients in SFA is set in a range of 10 to 130 with a step of 10. Normalisation of Fourier transformation is a parameter that is set for every dataset based on the training data. All datasets are standardised  before applying SAX. The strategy in SAX is to define the word length uniformly, which means all segments have identical \textcolor{black}{width}. The intervals for the bins are determined by minimum and maximum of the input data. 

Time series classification, as many classification tasks, is prone to poor accuracy due to class imbalance. Therefore, we apply an oversampling technique to pre-process imbalanced training data when exists. \textcolor{black}{Synthetic Minority Over-sampling TEchnique (SMOTE) \cite{chawla2002smote} demonstrated a good performance in oversampling of sample sets, whenever imbalance exists. It randomly creates and generates new minority class samples based on a certain rule and adds these newly synthesised samples into the original dataset to generate new training instances.} We will discuss in the following section how oversampling of imbalance datasets affected the overall accuracy, and whether it caused any overfitting/underfitting due to the amount of synthetically generated data. 
To facilitate extension of this work, and also for reproducibility, we will made the results and code available online \footnote{https://github.com/zabdallah/Co-eye}. 

\textcolor{black}{When applied to balanced datasets, any classifier is typically evaluated by predictive accuracy which is defined as the number of correctly classified instances divided by the total number of instances. When evaluating Co-eye accuracy, we use the standard accuracy/error measures to be consistent with all other experiments reported in the literature according to \cite{UCR}. However, predictive accuracy might not be appropriate when the data is imbalanced \cite{chawla2009data}. The main goal for learning from imbalanced data is to improve the recall without impacting the precision. Following this strategy, as an exception, we use precision/recall for measuring the impact of oversampling technique on imbalanced datasets (Section \ref{hyperparam2}).}

\subsection{Analysis of parameters}
\label{parameters}
\textcolor{black}{We analyse and justify  in this section the selection of parameters and techniques in Co-eye. We first investigate the number of lenses that are automatically generated via $SearchLenses$, discussed in Algorithm \ref{paramAlg}, in Section \ref{hyperparam1}. Then, we demonstrate the impact of applying SMOTE with the existence of class imbalance in Section in Section \ref{hyperparam2}. Section \ref{hyperparam3} discusses the advantage of combining both representations of time and frequency domains in Co-eye. The strategy of selecting lenses is then discussed in Section \ref{hyperparam4}. Finally, we report how other base classifiers perform with Co-eye compared to Random Forest in Section \ref{hyperparam5}.}

\subsubsection{\color{black}Number of lenses}
\label{hyperparam1}
\begin{figure}[h!]
    \centering
    \includegraphics[width=0.6\textwidth]{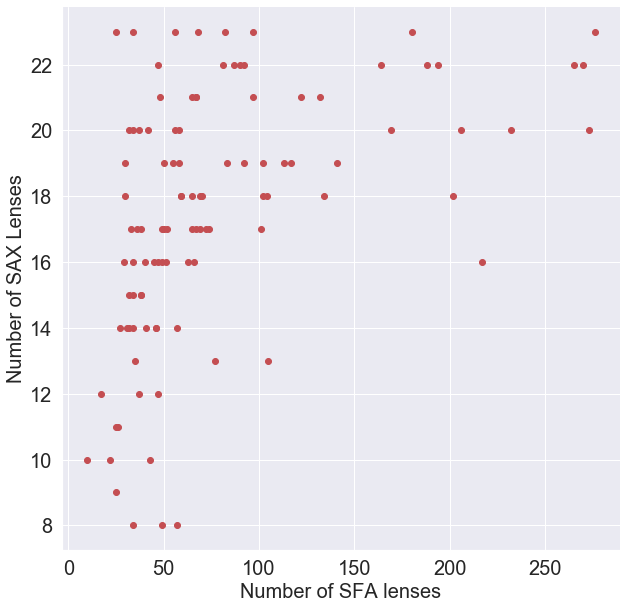}
      \centering
    \caption{No of lenses generated by SAX and SFA for each dataset}
    \label{pairs}
\end{figure}

We first evaluate the selection of lenses for both SAX and SFA. Figure \ref{pairs} shows the number of selected lenses in each, SFA on the x-axis and SAX on the y-axis. Each dot represents a dataset in the UCR collection. The graph shows the wide range of selections which mainly vary in the range of 100 to 250 with some exceptions. The correlation between the number of SAX pairs and their corresponding SFA pairs is non-linear, which indicates the diversity of selected representations as they are examining different perspectives of the data. As the SAX word length is selected uniformly in the experiments, the number of lenses in SAX reflects only the \textcolor{black}{size} of alphabets used. For example, ``Handsoutline'' dataset is one of the longest series of length 2709 with 2 classes. The number of SAX pairs is 22 while SFA pairs are 194. This suggests the diversity in both frequency-domain and time-domain for this dataset. The number of lenses observed in ``BME'' dataset for SAX is 8 while SFA pairs are 50. This suggests that fluctuation in the frequency domain is more significant in this dataset.  

\begin{figure}[!h]\centering
\begin{subfigure}{.7\textwidth}
 
  \includegraphics[width=\linewidth]{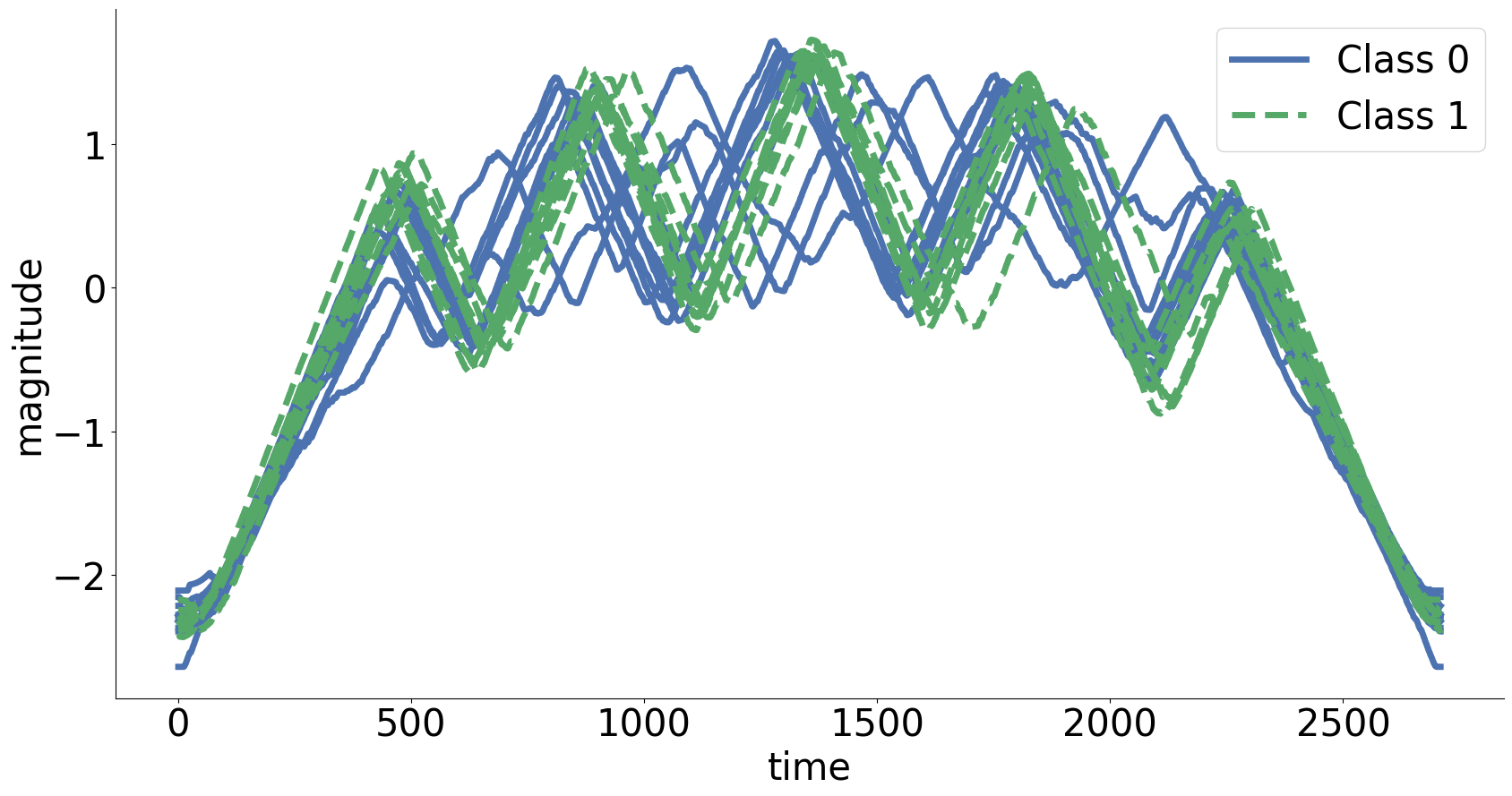}
 \caption{\color{black} An example of data that is diverse in both frequency and time domains}
   \label{HandOutlines_sub}

\end{subfigure}
\\

\begin{subfigure}{.7\textwidth}
  \centering
  \includegraphics[width= \linewidth]{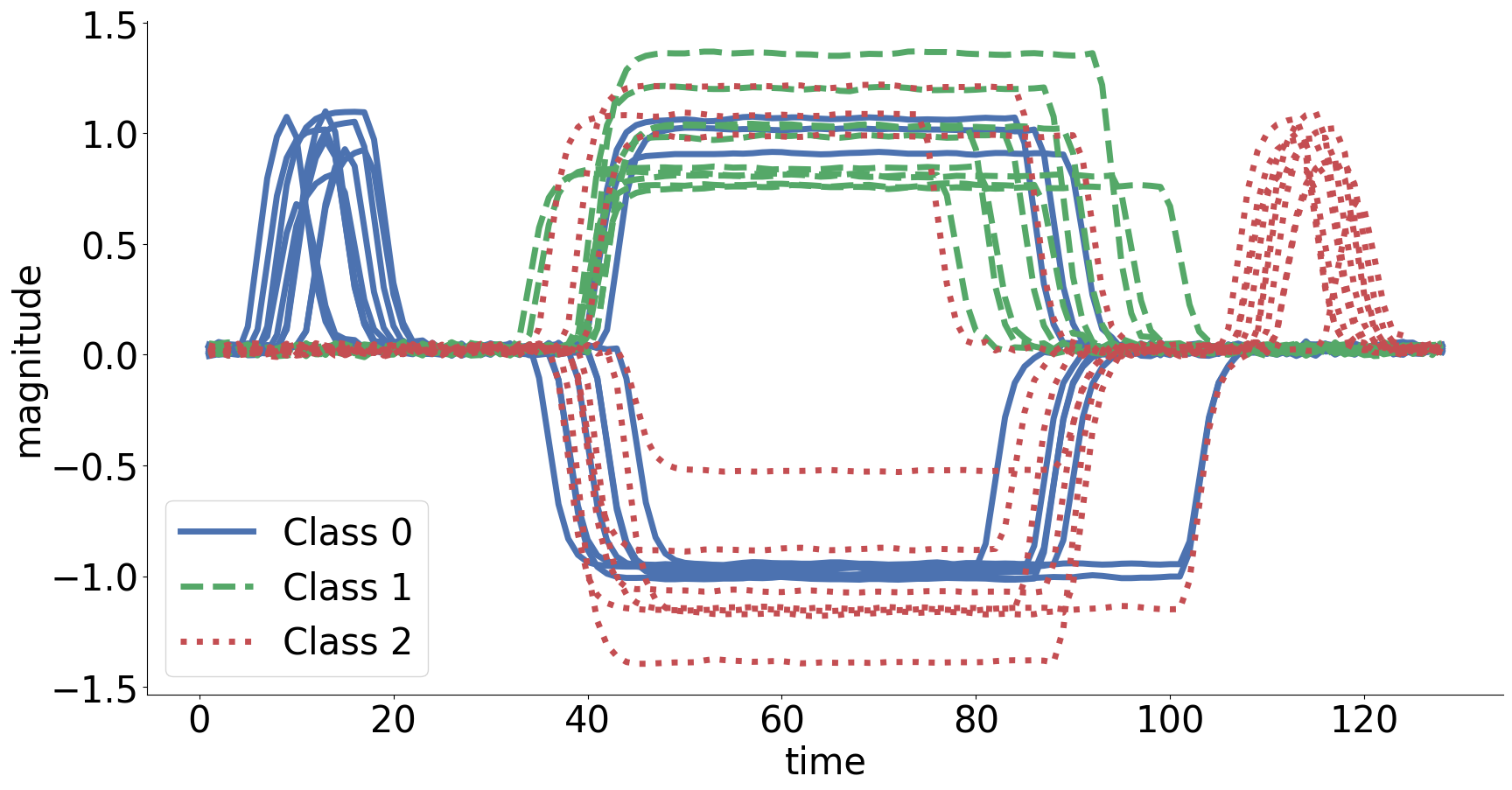}
   \caption{\color{black} An example of data that is diverse in frequency domain more than time domain}
     \label{BME_sub}

\end{subfigure}
\caption{\color{black}Samples of different classes in HandsOutlines and BME datasets}
\label{SAXSFA}
\end{figure}

We confirmed these observations by visually examining both datasets. As shown in Figure \ref{SAXSFA}, ``Handsoutlines'' fluctuation is in both time and frequency domains \ref{HandOutlines_sub}, while ``BME'' has a wide range of frequencies \ref{BME_sub}. It is also noted that the number of classes and the length of series have no direct correlation with the number of pairs generated. Thus, the parameter selection procedure is only governed by the diversification in both domains, time and frequency. 

It is worth noting that the hyper-parameterisation for lenses selection is a parameter-free algorithm that only requires the training data as an input. All variables that are used internally in the algorithm are data-driven and require no previous setup. For instance, $Threshold$ in line 8 in Algorithm \ref{paramAlg} is automatically generated based on the best accuracy of all lenses produced from cross-validation on the training set. 

\subsubsection{\color{black}Dealing with class imbalance}
\label{hyperparam2}
We also analyse how SMOTE impacted the overall accuracy across imbalance datasets. \textcolor{black}{Class imbalance is observed whereas a non-equal distribution of samples among classes exists.} SMOTE is only applied on training datasets that has imbalance distribution of classes in condition that the class has more than one instance, total of 70 datasets. Co-eye accuracy has been measured with and without SMOTE. We found that the difference ranges from 25\% gain to -4\% loss in accuracy, with an average of 4\% increase. The impact on accuracy is correlated to the percentage of increase in data samples due to oversampling. \textcolor{black}{SMOTE percentage indicates the percentage increase of the number of samples added via oversampling proportional to the original sample size}. 100\% increase means the dataset has been doubled to attain balance amongst classes.

\textcolor{black}{Table \ref{SMOTEacc} reports precision(P), recall(R) and F-measure(F1) of Co-eye on a subset of UCR datasets which has 20\% or more imbalance percentage. The first part of the table, before the line, represents datasets with binary/two classes. The rest of the table are datasets with more than two classes. Applying SMOTE improves the recall percentage in all cases (binary or multi-classes). F-measure shows an increase of 3\% in binary classification, 5\% in multi-class datasets and overall. In addition to the overall accuracy boost with SMOTE, the results also show that 31 datasets (out of 45) have better accuracy, in terms of F-measure, when SMOTE is applied. It is also shown that SMOTE has a better impact on multi-class datasets than binary ones. }

\begin{table}[h!]
\caption{Co-eye accuracy measures with and without SMOTE in imbalanced datasets. Datasets with binary classes are above the line, while multi-class datasets are below the line. P: precision, R: Recall, F1: F-measure}
\begin{tabular}{clp{1cm}||lls|lls}\\ \hline
                           & \multirow{2}{4em}{Dataset} & \multirow{2}{4em}{SMOTE Perc} & \multicolumn{3}{c|}{Co-eye with SMOTE} & \multicolumn{3}{c}{Co-eye without SMOTE} \\ \cline{4-9}
                             &&&  P    & R   & F1  & P     & R& F1    \\ \hline

\multirow{13}{*}{\rotatebox[origin=c]{90}{Binary}}&DistalPhaOutlCorrect  &26\%  & 0.74         & 0.74     &\textbf{0.74 }       & 0.75          & 0.74      & \textbf{0.74 }        \\

&Wafer                    &81\%      & 0.95         & 0.99     & 0.97        & 0.99          & 0.98      &\textbf{ 0.99}         \\

&MiddlePhaOutlCorrect  &29\%  & 0.73         & 0.72     & 0.71        & 0.75          & 0.75      & \textbf{0.75}         \\

&Earthquakes        &64\%            & 0.88         & 0.51     &\textbf{0.46 }       & 0.37          & 0.50      & 0.43         \\

&ProximalPhaOutlCorrect &35\% & 0.87         & 0.86     &\textbf{0.86}        & 0.88          & 0.78      & 0.81         \\

&SonyAIBORobotSurface1  &40\%        & 0.87         & 0.88     &\textbf{0.87}        & 0.76          & 0.67      & 0.60         \\

&ECG200   &38\%                      & 0.86         & 0.88     & 0.86        & 0.87          & 0.86      &\textbf{ 0.87 }        \\

&Lightning2           &33\%          & 0.77         & 0.77     &\textbf{ 0.77 }       & 0.75          & 0.70      & 0.70         \\

&PhalangesOutlinesCorrect   &30\%    & 0.76         & 0.77     & 0.76        & 0.79          & 0.77      & \textbf{0.78}         \\

&Strawberry           &29\%          & 0.92         & 0.94     & 0.93        & 0.94          & 0.95      & \textbf{0.94}         \\

&HandOutlines       &28\%            & 0.89         & 0.89     & 0.89        & 0.91          & 0.90      &\textbf{ 0.90 }        \\

&ECGFiveDays      &22\%              & 0.86         & 0.86     & \textbf{0.86 }       & 0.82          & 0.81      & 0.81         \\

&Herring        &21\%                & 0.51         & 0.51     & \textbf{0.51}        & 0.48          & 0.48      & 0.47         \\ \hline
\multirow{30}{*}{\rotatebox[origin=c]{90}{Multi-class}}&WordSynonyms               &462\%    & 0.48         & 0.46     &\textbf{ 0.45 }       & 0.49          & 0.36      & 0.38         \\
&MedicalImages               &433\%   & 0.58         & 0.73     & 0.63        & 0.78          & 0.60      & \textbf{0.65 }        \\
&DistalPhalanxTW                &170\%& 0.50         & 0.48     & \textbf{0.49 }       & 0.35          & 0.38      & 0.33         \\
&ECG5000                     &192\%   & 0.67         & 0.57     &\textbf{ 0.60 }       & 0.69          & 0.50      & 0.54         \\
&ProximalPhalanxTW           &170\%   & 0.41         & 0.43     &\textbf{0.41}        & 0.39          & 0.43      & \textbf{0.41}         \\
&MiddlePhalanxTW         &141\%       & 0.37         & 0.38     &\textbf{0.38}        & 0.36          & 0.38      & 0.37         \\
&FacesUCR                 &131\%      & 0.81         & 0.79     &\textbf{0.80}        & 0.83          & 0.73      & 0.76         \\
&Worms                      &110\%    & 0.52         & 0.49     &\textbf{0.50}        & 0.64          & 0.42      & 0.45         \\
&Symbols                     & 92\%  & 0.91         & 0.89     &\textbf{0.89}        & 0.85          & 0.83      & 0.82         \\
&Lightning7                   &90\%  & 0.60         & 0.63     &\textbf{0.59}        & 0.60          & 0.65      &\textbf{ 0.59}         \\
&OliveOil                     &73\%  & 0.85         & 0.82     &\textbf{ 0.82 }       & 0.92          & 0.79      & 0.79         \\
&StarLightCurves               &72\% & 0.94         & 0.96     &\textbf{0.95}        & 0.97          & 0.94      & \textbf{0.95}         \\
&ChlorineConcentration      &   68\% & 0.66         & 0.65     &\textbf{0.65}        & 0.72          & 0.58      & 0.61         \\
&Mallat                      &  60\% & 0.96         & 0.96     &\textbf{0.96}        & 0.93          & 0.91      & 0.91         \\
&OSULeaf                      & 59\% & 0.60         & 0.61     &\textbf{ 0.59 }       & 0.58          & 0.52      & 0.49         \\
&InsectEPGRegularTrain          &45\%& 0.81         & 0.84     &\textbf{0.82 }       & 0.83          & 0.75      & 0.76         \\
&Adiac                         & 42\%& 0.75         & 0.76     &\textbf{0.74 }       & 0.69          & 0.73      & 0.69         \\
&InsectEPGSmallTrain          & 41\%  & 0.74         & 0.74     &\textbf{0.74  }      & 0.89          & 0.66      & 0.66         \\
&ProxPhalxOutlAgeGroup & 42\% & 0.76         & 0.82     & 0.78        & 0.77          & 0.80      & \textbf{0.78 }        \\
&FaceFour                 &   33\%   & 0.85         & 0.87     &\textbf{ 0.85 }       & 0.69          & 0.63      & 0.59         \\
&Plane                     & 33\%    & 0.97         & 0.97     & \textbf{0.97 }       & 0.97          & 0.97      & \textbf{0.97}         \\
&NonInvasFetECGThorax1   &31\%  & 0.89         & 0.90     &\textbf{0.89}        & 0.89          & 0.89      & 0.89         \\
&NonInvasFetECGThorax2  & 31\%  & 0.92         & 0.92     &\textbf{0.92}        & 0.92          & 0.92      & \textbf{0.92}         \\
&CinCECGTorso         &30\%         & 0.80         & 0.80     &\textbf{0.80}        & 0.75          & 0.73      & 0.71         \\
&MidPhalOutlAgeGroup  &78\% & 0.51         & 0.48     &\textbf{0.49}        & 0.45          & 0.39      & 0.34         \\
&CricketZ            &26\%           & 0.58         & 0.60     & 0.59        & 0.61          & 0.62      &\textbf{ 0.61}         \\
&DisPhalxOutlAgeGroup  &93\%&  0.67         & 0.72     &\textbf{0.66}        & 0.63          & 0.53      & 0.54         \\
&InlineSkate   &26\%                 & 0.33         & 0.33     &\textbf{ 0.32 }       & 0.34          & 0.30      & 0.31         \\
&SwedishLeaf        &26\%            & 0.91         & 0.91     &\textbf{0.91}        & 0.89          & 0.89      & 0.89         \\
&Trace         &24\%                & 0.98         & 0.98     & 0.98        & 0.99          & 0.99      & \textbf{0.99}         \\ \hline \hline
\multicolumn{3}{c||}{\textbf{ Mean (Binary-classes)} }          & \textbf{0.82 }        & \textbf{0.79}     & \textbf{0.78}        & 0.77          & 0.76      & 0.75     \\
 \multicolumn{3}{c||}{\textbf{Mean (Multi- classes})}        & \textbf{0.71 }        & \textbf{0.72}     & \textbf{0.71}        &\textbf{ 0.71}         & 0.66     & 0.66     \\  
\multicolumn{3}{c||}{ \textbf{ Mean (All)} }       & \textbf{0.74 }        & \textbf{0.74}     & \textbf{0.73}        & 0.73          & 0.69      & 0.69     \\\hline   
\end{tabular}
\label{SMOTEacc}
\end{table}

\textcolor{black}{The aforementioned results showed that balancing the dataset generally has a positive impact on accuracy, with more than 10\% accuracy improvement in multiple cases. An extreme overfitting, such as ``MedicalImages'' dataset that has more than 400\% increase in data, might cause an accuracy loss, 2\% in this case. In numerous cases when datasets are complemented with a very small percentage for balancing, SMOTE has no/slight impact on accuracy. Balancing attained a remarkable accuracy improvement in ``SonyAIBORobotSurface1'' with 27\% accuracy increase and ``MiddlePhalanxOutlineAgeGroup'' with 15\%  accuracy increase. Throughout the experiments, we use SMOTE by default in the pre-processing step for balancing imbalanced data. }

\subsubsection{\color{black}Combination of both representations}
\label{hyperparam3}
\begin{figure}[h!]
    \centering
    \includegraphics[width=\textwidth]{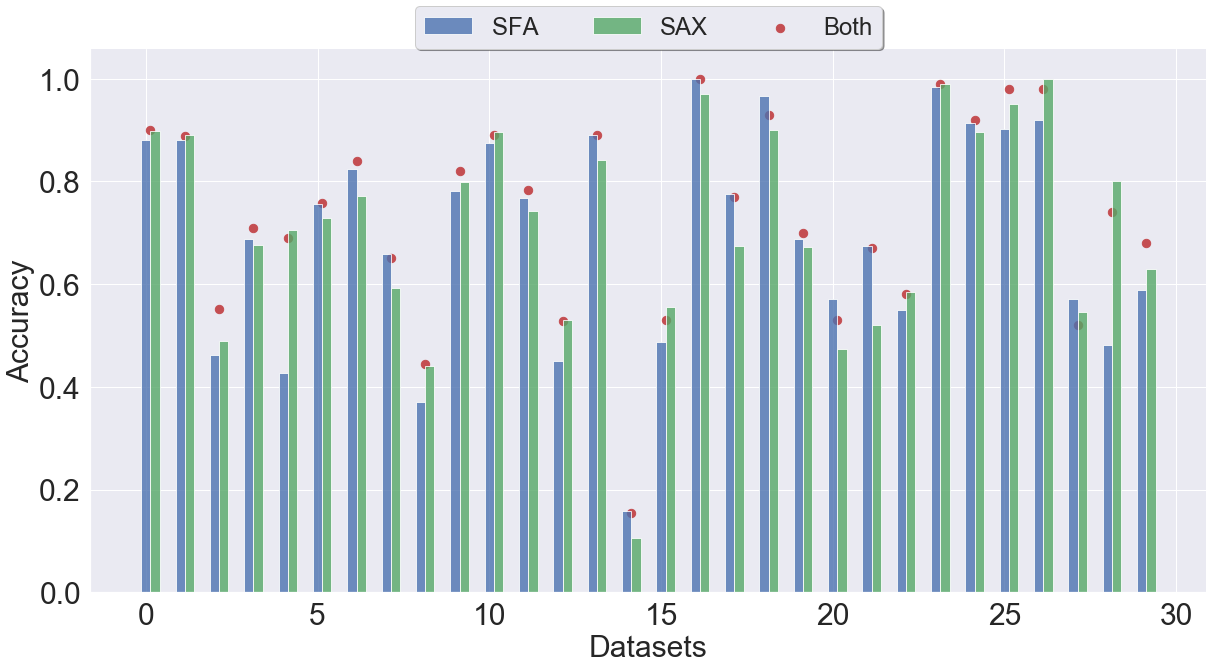}
      \centering
    \caption{Co-eye with SAX only, SFA only and both}
    \label{SAXSFAfig}
\end{figure}

We also investigate the value of combining eyes from both representations (SAX and SFA). \textcolor{black}{In this experiment, we implement Co-eye algorithm, however, in voting, we consider either only SAX, only SFA or a combination of both using the aforementioned voting mechanism. To be able to visualise the difference, we choose 30 random datasets to display the results. Figure \ref{SAXSFAfig} shows the accuracy of Co-eye using only an individual symbolic representation in addition to Co-eye accuracy using both representations across the randomly selected datasets.} For each representation, the most confident lens is selected. The black bar represents SFA only accuracy, while the green bar is SAX only accuracy. Co-eye, with the red dot, moderated between the most confident selection from each representation to choose the predicted label. It is clear from \textcolor{black}{this figure} that the voting mechanism chooses the best from the two representations, with Co-eye that combines both achieves the best, or more, out of both.

\color{black}
\subsubsection{Lenses selection strategy }
\label{hyperparam4}

\begin{figure}[!h]
    \centering
    \includegraphics[width=\textwidth]{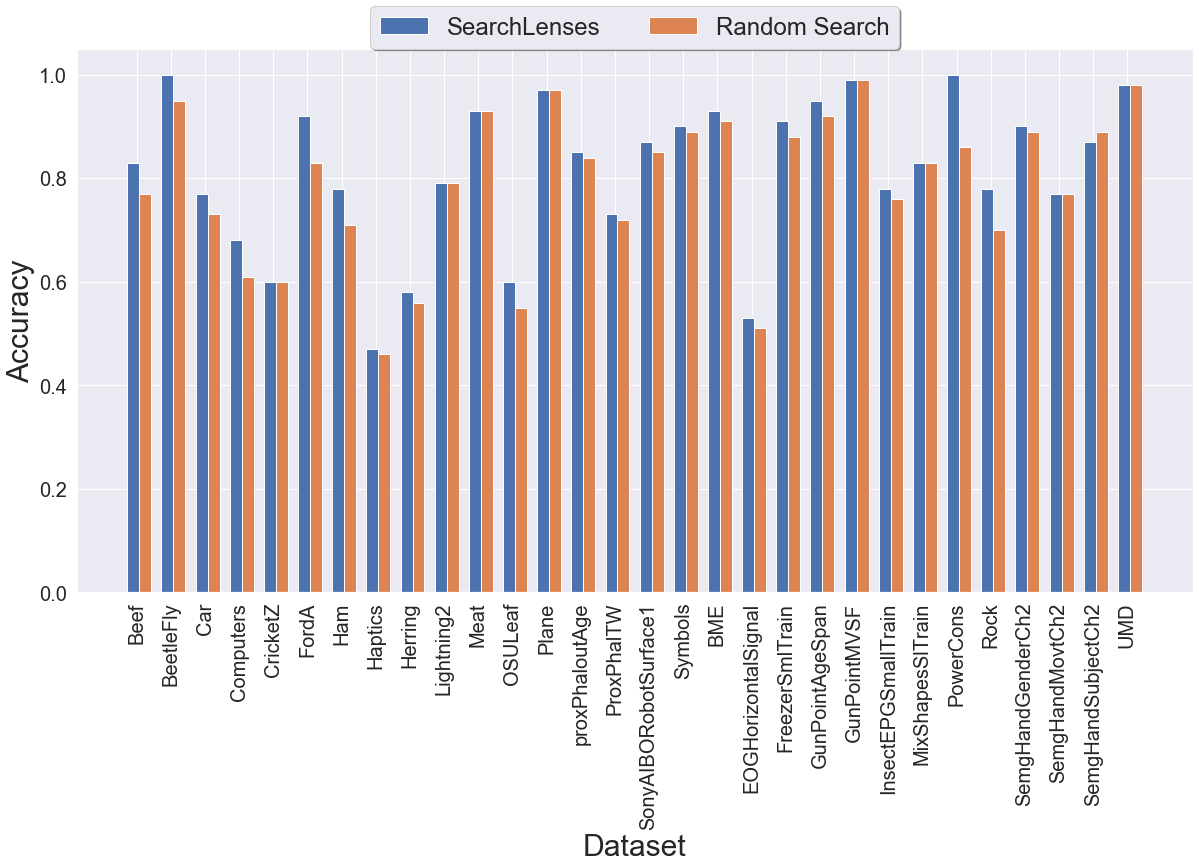}
    \caption{Impact of SearchLenses and Random Search on Co-eye accuracy }
    \label{randomSearch}
\end{figure}

In this section we investigate the strategy of lens selection in Co-eye. SearchLenses Algorithm presented in Section \ref{hyper-param} discussed in details the mechanism Co-eye uses to search regions of sharp/accurate lenses. It is noted from the literature that random search also performs well for hyper-parameterisation \cite{bergstra2012random}. Hence, we compare the performance of Co-eye using both strategies, SearchLenses and random search. The analysis is performed on 30 random datasets of the UCR archive. Figure \ref{randomSearch} shows that applying SearchLenses strategy achieves either more or equally accurate results compared to random search, with only one exception (SemgHandSubjectCh2 dataset). The improvement is recorded the best in PowerCons dataset with 14\% increase, 9\% in FordA dataset, and 8\% in Rock dataset. 

\subsubsection{Co-eye base classifier}
\label{hyperparam5}
Random Forest is implemented in the classification phase for each lens in Co-eye. Then, Co-eye implements voting between the most confident random forests in each presentation to choose the best lenses that best fit a specific dataset.  To validate the superiority of Random Forest in Co-eye, we evaluated Co-eye performance with other base classifiers too. Thus, Random Forest is replaced by the other classifiers to run this experiments for each dataset. Best parameters are used in each classifier. Parameters are as follow: 
\begin{itemize}
    \item Support Vector Machine (SVM): kernels are Gaussian, Gamma= $1 / no. features$, Regularisation parameters is set to a high value (1e6). The strength of the regularisation is inversely proportional to this value.
    \item Rotation Forest (RotForest): an ensemble of 100 forests, using the Gini coefficient. 
    \item Gradient Boosting (Gradient Boosting): with 100 estimators and learning rate of 1.0. 
    \item AdaBoost: with 100 estimators and learning rate of 1.0. 
\end{itemize}

\begin{figure}[!h]
    \centering
    \includegraphics[width=\textwidth]{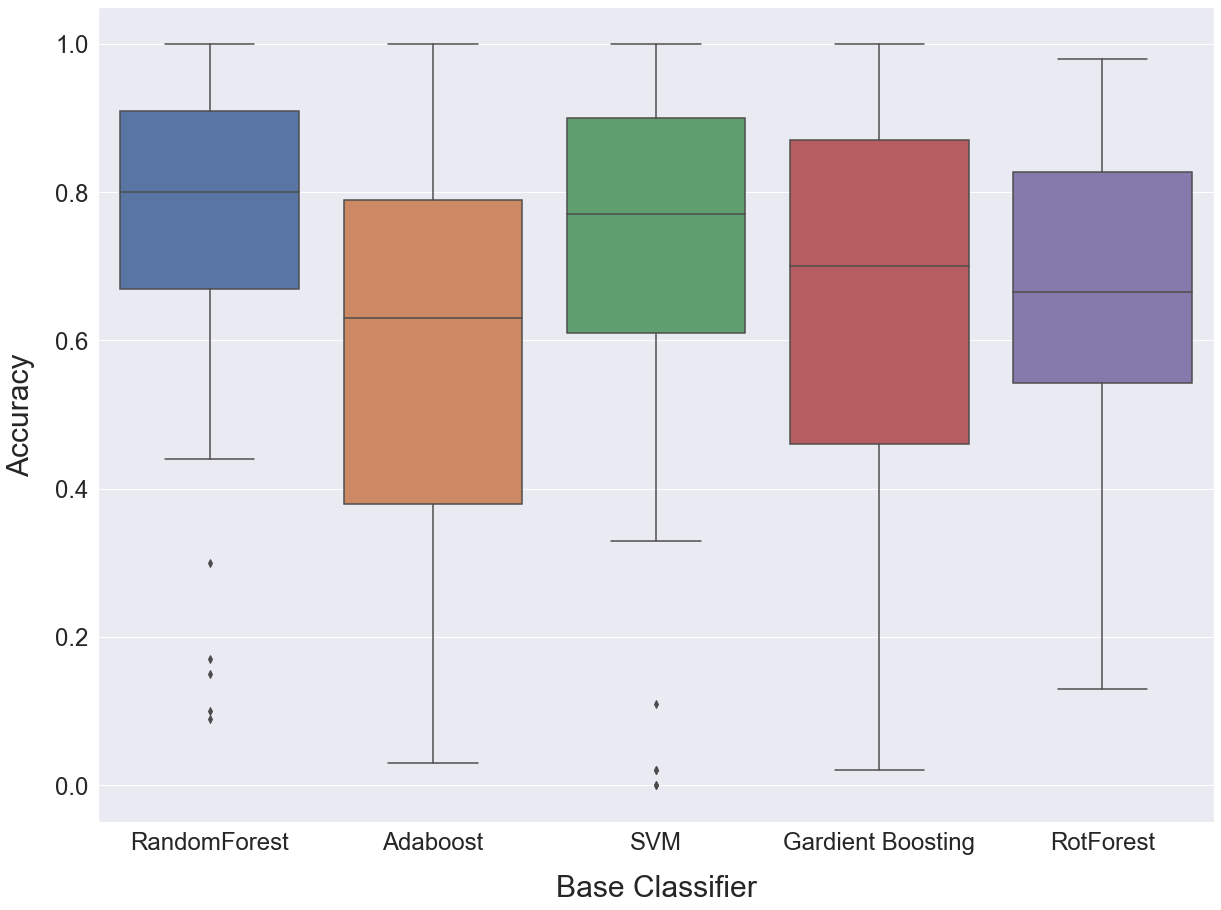}
    \caption{Co-eye with various base classifier}
    \label{baseClf}
\end{figure}

Figure \ref{baseClf} shows Co-eye accuracy variation across the 114 datasets with various base classifiers. Random Forest has the highest mean accuracy across all. It is also notable that interquartile range of the box plot (IQR), which is simply its width, is the smallest in Random Forest compared to other classifiers. IQR reflects the spread of accuracy around the mean value, which is better in Random Forest than other base classifiers. SVM comes next in terms of mean accuracy and spread. However, SVM has completely failed to find decision boundaries in some of the datasets (appeared as outliers in the box plot). One of these datasets is ``Fungi''  that has 0\% accuracy using SVM, compared to 84\% when using Random Forest.

From the above, we can conclude that Random Forest is the most suitable base classifier for Co-eye, typically because Random Forest possesses two features that are coherent with Co-eye mechanism and objectives which are diversity and robustness to overfitting.
\color{black}

\subsection{Analysis of classification accuracy}
\label{co-eye_acc}

\color{black}{
After discussing variations of parameters in Co-eye, in this section, we evaluate Co-eye performance in comparison with other classification methods. Co-eye in the following implements Random Forest as a base classifier, applies SMOTE for imbalance data, and combines both SFA and SAX representations. We evaluate Co-eye performance on the benchmarked UCR repository \cite{UCR}. An extended version of UCR datasets has been released recently with 128 datasets, of which 114 datasets with non-varied lengths. In order to be consistent with the published results in \cite{UCR,TSCwebsite, bagnall2017great}, we follow the same train/test split and same performance measures. There are two sets of published experiments:
\begin{itemize}
    \item Set 1: UCR repository \cite{UCR} which contains the newly published datasets with the classification accuracy reported for three benchmark classifiers: \textcolor{black}{Euclidean Distance (ED) $k$-nearest neighbour with $k=1$ , Dynamic Time Warping (DTW) with a fixed window of 100 and DTW with a learned window.}
    \item Set 2: Bangnall et al \cite{TSCwebsite, bagnall2017great} have recently published a survey that reports a comprehensive analysis of many TSC algorithms on the benchmark of 85 datasets (the old UCR repository).
\end{itemize}}
We assess the performance of Co-eye on both sets. We first perform a pairwise comparison of Co-eye performance with the published results of the new repository (Set 1) using the three benchmark classifiers: ED, DTW with a fixed window and DTW with a learned window in Section \ref{pairwis}. The new repository also offers a wide range of domains, hence, we evaluate Co-eye performance across domain on the same new repository (Set 1) in  Section \ref{domains}. Then, we compare Co-eye with a wide range of methods presented in Set 2 in Section \ref{allmethodsSec} .We finally discuss Co-eye time complexity in Section \ref{time}.

\color{black}
\subsubsection{Pairwise comparison}
\label{pairwis}
\begin{figure}[!h]
    \centering
    \includegraphics[width=0.7\textwidth]{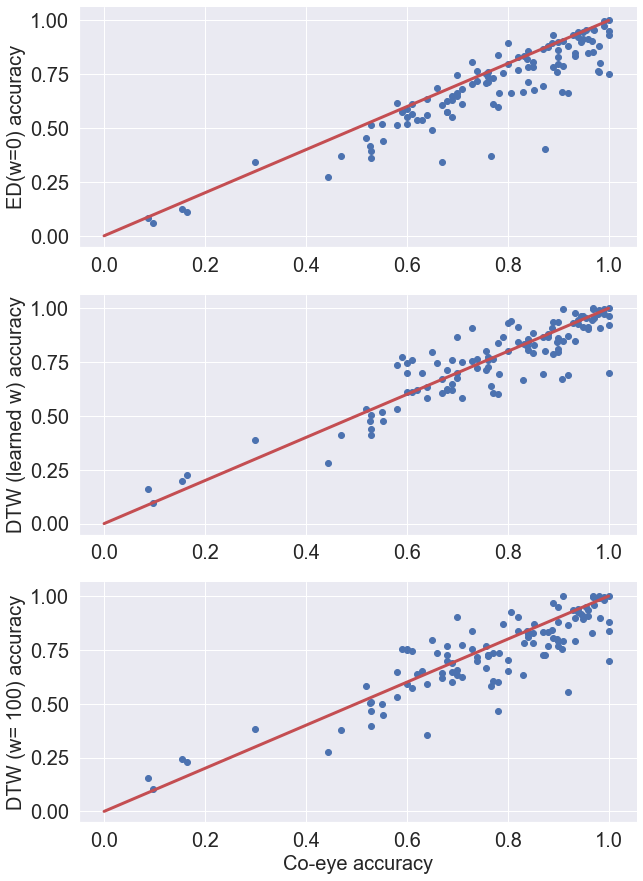}
    \caption{Pairwise comparison of Co-eye against benchmark classifiers ED, DTW(w=100) and DTW (learned w) on 114 UCR datasets}
    \label{pairwise}
\end{figure}

The scatter plots in Figure \ref{pairwise} shows a pairwise comparison of the classification accuracy on test set. \textcolor{black}{The results for different methods are reported in \cite{TSCwebsite, bagnall2017great}}. Each dot represents a dataset. A dot below a line indicates that Co-eye outperforms the opponent classifier. More significant accuracy improvement is farther from the diagonal. The scatter plots show that Co-eye is better than ED for most datasets, 92 (more than 80\% of the datasets), with a tie in another 5. The most significant improvement is in the Spectrum domain datasets of ``SegHandSubjetCh2'',  ``SegHandMovementCh2'' and also power consumption dataset, ``smallKitchenAppliances''. Co-eye outperforms DTW with fixed window (w=100) in 72 datasets, with best improvement in datasets ``Ham'', ``InsectWingBeatSound'' and ``EthanolLevel''. In comparison to DTW with learned window, Co-eye shows an improvement in accuracy for 60 datasets. Datasets such as ``FordA'' shows the best improvement in accuracy, where $DTW_W$ attains an accuracy of 69\% while Co-eye's accuracy is 92\%. A similar improvement of 16\% achieved in a challenging spectrum dataset of ``EthanolLevel'', first introduced in \cite{lines2016hive}.

\subsubsection{Across-domains performance}
\label{domains}
The release of an expanded version of the UCR repository enabled us to perform extensive analysis on datasets from a diverse range of domains. Co-eye has a unique feature of bringing together different perspectives with the compound eye. Therefore, it is expected to have a robust performance across domains. Table \ref{compDomain} reports the mean classification error of datasets corresponding to each domain. Count refers to the number of datasets represented in the repository for each domain. \textcolor{black}{The last row represents the total number of winning domains for each classification method. If two or more methods equally achieve the lowest error, they both share the best rank. Co-eye has the lowest classification error for 8 different domains and share the first rank with three others. Hence, Co-eye has the best ranking, of 9.5, among other methods, while Dynamic Time Warping with a learned window comes next with only 2.5.} Co-eye performs best with Spectro and spectrum, with 5-6\% more accurate classification in both. Traffic domain is represented by two challenging datasets, ``Chinatown'' and ``MelbournePedestrains''. Co-eye attains the best accuracy on both datasets of 93\% and 97\% for ``Chinatown'' and ``MelbournePedestrains'', respectively. One important finding of these results is that Co-eye attains its best performance with datasets that have no-massive phase shifting. The mechanism of selecting lenses and generating random forests based on these lenses requires an approximate alignment. Enhancing the performance of Co-eye for series with phase-shifting is a priority for our future development. 
\begin{table}[h!]
\caption{\color{black}Mean classification error of each methods on UCR datasets grouped by domains (smaller is better). The last row indicates the total number of domains when the method ranked first (for each method)}
\begin{tabularx}{\textwidth}{cccccc}

\toprule
Type          & Count & ED   & $DTW_{w=100}$ & $DTW_{lW}$ & Co-eye \\ \midrule
Device        & 8     & 0.51 & \textbf{0.35}       & 0.36           & \textbf{0.35}  \\
\textbf{ECG}           & 6     & 0.16 & 0.16       & 0.14           & \textbf{0.13}  \\
\textbf{EOG}           & 2     & 0.57 & 0.52     & 0.52           & \textbf{0.46}  \\
\textbf{EPG}           & 2     & 0.33 & 0.20      & 0.24           & \textbf{0.18} \\
HRM           & 1     & 0.18 & \textbf{0.16}       & 0.18           & \textbf{0.16}  \\
Hemodynamics  & 3     & 0.91 & \textbf{0.83}       & 0.85           & 0.89  \\
Image         & 32    & 0.28 & 0.26       & \textbf{0.24}           & \textbf{0.24}  \\
Motion        & 17    & 0.31 & 0.27       & \textbf{0.25}           & 0.28  \\
\textbf{Power}         & 1     & 0.07 & 0.12       & 0.08           & \textbf{0.0}  \\
\textbf{Sensor}        & 20    & 0.27 & 0.25       & 0.22           & \textbf{0.19}  \\
Simulated   & 8     & 0.18 & 0.09       & \textbf{0.06}           & 0.08\\

\textbf{Spectro}       & 8     & 0.26 & 0.29       & 0.26           & \textbf{0.20}  \\
\textbf{Spectrum}      & 4     & 0.41 & 0.32       & 0.22           & \textbf{0.17}  \\
\textbf{Traffic}       & 2     & 0.10 & 0.13       & 0.10         & \textbf{0.05}  \\\bottomrule

Total number of winning in domains&      & 0& 1.5          & 2.5              & \textbf{9.5}     \\ \bottomrule
\end{tabularx}

\label{compDomain}
\end{table}

\subsubsection{Comparison with state-of-the-art TSC methods}
\label{allmethodsSec}
Bangall et al \cite{bagnall2017great} recently published a comprehensive analysis of many TSC algorithm on the benchmark of 85 datasets. We compare Co-eye with relevant state-of-the-art methods described in this survey. We consider a method as relevant if its classification model is based on trees or if it applies a symbolic representation transformation. We excluded some methods from the comparison for two main reasons. First, some methods fundamentally combine stand-alone methods for an accuracy boost. COTE, for instance, combines 35 classifiers into a single ensemble with a weighted vote. Therefore, we believe these methods provide a unique platform that brings powerful aspects of each individual classifier together. Co-eye is yet another stand-alone classifier that has its own strengths which we believe will contribute to enhancing the performance of these ensemble methods whenever it is integrated with. Second, some of these algorithms have a very long running time, such as ensembles of elastic distance measures (EE), which might also require special processing capabilities for a successful run, especially with the very long series. 

\begin{figure}[!h]
    \centering
    \includegraphics[width=\textwidth]{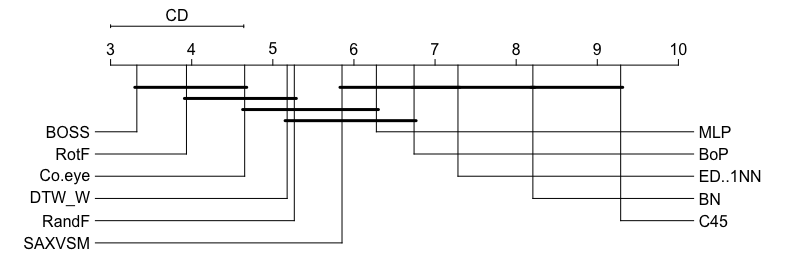}
    \caption {Critical difference (CD) diagram using Friedman Test and a post-hoc pairwise Nemenyi test comparing benchmark classifiers and Co-eye. High-to-low rankings run left to right. The higher the rank, the better the technique.}
    \label{allmethods}
    
\end{figure}

We also emphasise in these experiments on comparing Co-eye with SAXVSM as it is the closest relative technique to Co-eye amongst the current state-of-the-art techniques. According to experiments in \cite{bagnall2017great}, Rotation Forest \cite{rodriguez2006rotation} and DTW (with learned window) are considered as a benchmark for comparison based on an extensive analysis reported in this survey. Thus, we ensure both methods are used as a benchmark for our experiments too. We also included state-of-the-art BOSS \cite{schafer2015boss} technique as it uses a similar approach, however, it relies only on Fourier approximation and does not use lenses as Co-eye. We first plot the critical difference diagram following the same methodology described in \cite{demvsar2006statistical} when testing for significant difference among classifiers. Figure \ref{allmethods} depicts the significant difference in ranks among classifiers using Friedman Test and a post-hoc pairwise Nemenyi test. The diagram shows the average rank of classifiers, over 85 datasets, in order. The higher the rank, the better the technique. 
\textcolor{black}{The x axis where the lines end represents the average rank position of the respective methods across all datasets. The null hypothesis is that the average ranks of each pair of methods do not differ with statistical significance. Horizontal lines connect the lines of the methods for which we cannot exclude the hypothesis that their average ranks are equal. Any pair of methods whose lines are not connected with a horizontal line can be seen as having an average rank that is different with statistical significance. Hence, in Figure 15, although C4.5 (of rank 9.3), and SAXVSM (of rank 5.8) are different in terms of ranking, the difference is not statistically significant (according to  Nemenyi test). The opposite  is also valid; the difference in ranking between two methods can be small, yet the difference can be statistically significant (such as BOSS and $DTW_w$)} 
Among 11 other classifiers, Co-eye is ranked third compared to other state-of-the-art techniques following BOSS and Rotation Forest. Yet, Co-eye average rank is higher than the other benchmark (DTW with learned window). Co-eye rank is also higher than Random Forest, which suggests that applying Random Forest on the whole series is less accurate than using Random Forest through the concept of lenses introduced in Co-eye.  

\begin{figure}[!htp]
\centering
\begin{subfigure}{.48\textwidth}
  \centering
  \includegraphics[width=\linewidth]{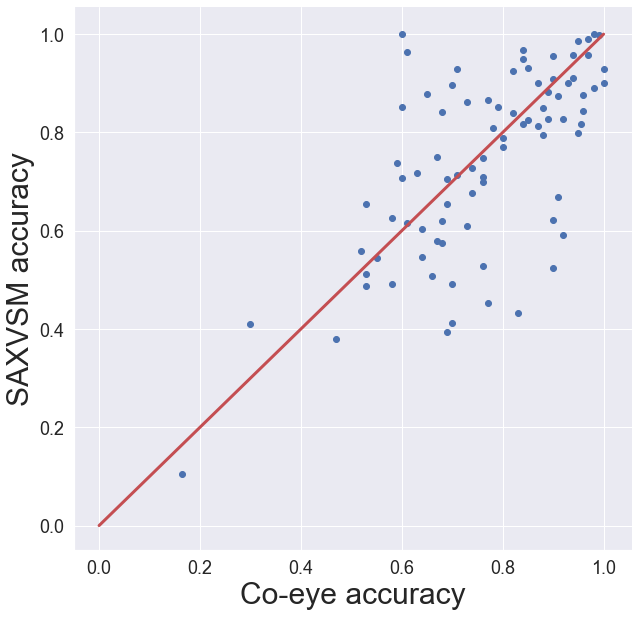}
  \caption{Co-eye vs. SAXSVM}
  \label{SAXVSM}
\end{subfigure}
\begin{subfigure}{.48\textwidth}
  \centering
  \includegraphics[width= \linewidth]{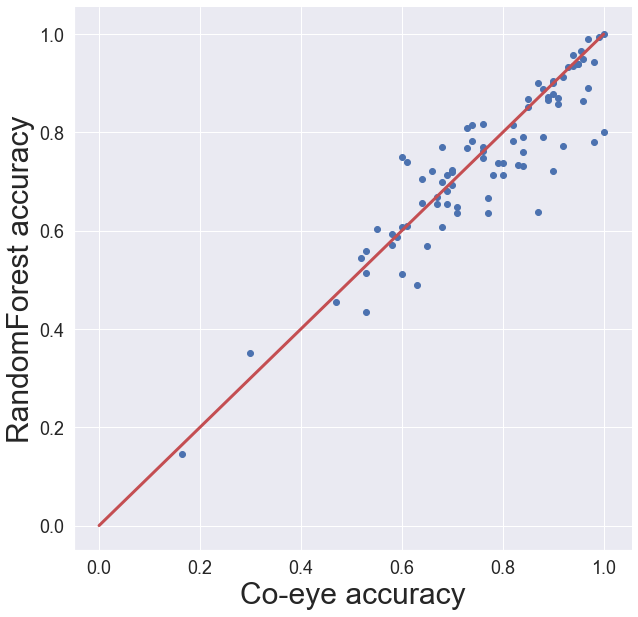}
  \caption{Co-eye vs. Random Forest }
  \label{RF}
\end{subfigure}
\begin{subfigure}{.48\textwidth}
  \centering
  \includegraphics[width=\linewidth]{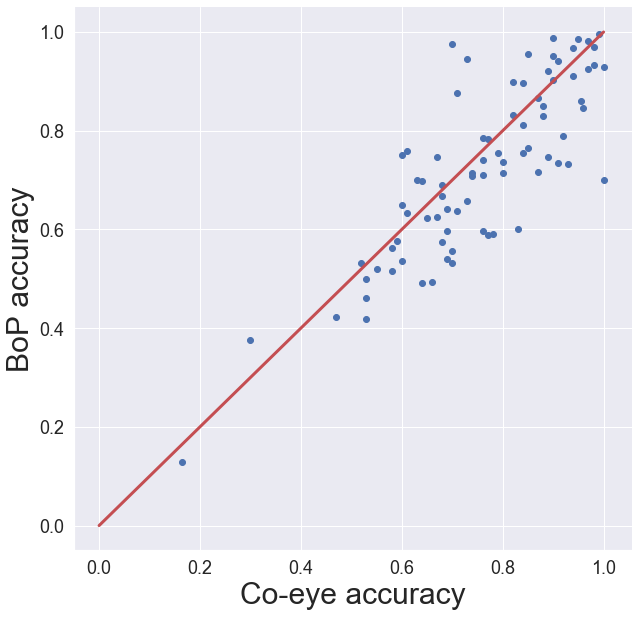}
  \caption{Co-eye vs. BoP}
  \label{BoP}
\end{subfigure}
\begin{subfigure}{.48\textwidth}
  \centering
  \includegraphics[width= \linewidth]{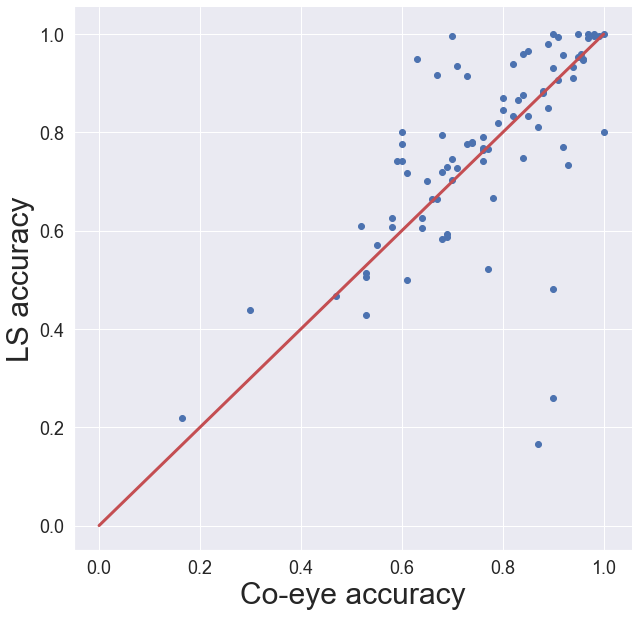}
  \caption{Co-eye vs. LS }
  \label{LS}
\end{subfigure}

\caption{Pairwise comparison of Co-eye versus other relevant techniques on 85 UCR datasets}
\label{pairwise2}
\end{figure}

Figure \ref{pairwise2} displays a pairwise comparison between Co-eye and relevant methods. We evaluate the performance of Co-eye against SAXVSM as it has similarities with Co-eye in terms of usage of symbolic representation.

The results show an improvement of Co-eye performance for 50 datasets (out of 85). Some of these improvements are substantial, such as in ``Adiac'' dataset, SAXVSM reported an accuracy of 42.5\% while Co-eye's accuracy is 77\%.  Another example is ``Beef'' spectro dataset with an accuracy jump from  43.3\% of SAXVSM to 83.3\%. Two ECG datasets ``NonInvasiveFatalECGThorax1'' and ``NonInvasiveFatalECGThorax2'' show 37.61\% and 32.8\% enhancement in accuracy of Co-eye over SAXVSM. This is consistent with our findings, discussed earlier in this section, that Co-eye's main strengths are demonstrated with datasets with no significant phase shifting such as ECG and spectrum series. We also carried out pairwise comparison with Random Forest as it is the core classifier in Co-eye. We have reported that Co-eye outperforms Random Forest on 49 datasets. That confirms that the combination of forests and presentations in Co-eye contributes to more accurate classification across domains. BoP is a standard method in the literature for symbolic representation. The results show that Co-eye is more accurate than BoP in 53 datasets. Again, an improvement is reported in spectro datasets: Beef, Meat, Ham with an accuracy boost of 23.3\%, 19.9\% and 12.3\% respectively. Finally, we conducted the analysis on Learned shapelet (LS) \cite{grabocka2014learning} that is considered one of the  best-ranked method in TSC using shaplets \cite{bagnall2017great}. Although LS outperforms Co-eye in 53 datasets, Co-eye improvement is notable, when it wins. For example, in ``OliveOil'', LS reported accuracy is only 16.7\% while Co-eye accuracy is 87\%. Similarly, other spectrum datasets of ``Meat'' and ``Ham'' where Co-eye accuracy is 93\% and 78\% respectively with 19.6\% and 11.3\% improvement in accuracy compared to LS. A significant improvement is reported in ECG detests `NonInvasiveFatalECGThorax1' of 64\% improvement and `NonInvasiveFatalECGThorax2' with 14.9\%. The power datasets contain a very diverse time and frequency characteristics as they record appliances power consumption which includes a wide range of devices that follow various usage patterns. Although shaplets is expected to enable the discovery of these patterns of devices' usage, Co-eye performs better in this domain as it considers multi-resolutions and diversification across time and frequency domains. A complete list of results for Co-eye and other methods is reported in the appendix. 

\subsubsection{Co-eye complexity}
\label{time}
In terms of complexity, the bottleneck in Co-eye is in the hyper-parameterisation step in the training phase when cross-validation is performed in order to select the best lenses. Once the selection of lenses is completed, the classification phase requires only a symbolic transformation for TS to SAX and SFA for $k$ times where $k$= $N$+$M$, $N$ is the number of SAX pairs and $M$ is the number of SFA pairs (lines 2 and 6 in Algorithm \ref{classification}). PAA in SAX has a linear complexity in the length of the time series $\mathcal{O}(n)$. The transformation of SFA words of length $w$ over an alphabet of size $\alpha$ from a set of $S$ time series of length $n$ has a complexity of $\mathcal{O}(S. n \log{n})$. The MCB step in SFA training adds to the complexity as it creates a look-up table that is computed from the training set. Co-eye voting procedure is constant and requires small/insignificant running time.  
\color{black}

We report Co-eye running time for a set of datasets with various characteristics in terms of length, training and testing sizes. All experiments are performed on a machine with a processor of 2.3 GHz Intel Core i5 and 8GB RAM. As discussed, the main bottleneck in terms of running time in Co-eye is the hyper-parameterisation step. Hence, we report the total running time (in seconds) as well as time for hyper-parameterisation (SAX and SFA), training time following the selection of parameters and prediction time on test data.  Table \ref{tt} depicts the running time for each phase on the selected dataset. It is clear from this table that there are two main characteristics which control the running time, number of training series and the series length. The total time in the table is the time spent to run Co-eye end-to-end including hyper-parameterisation, training and prediction time. It is noted that hyper-parameterisation time is the most time-consuming step in Co-eye, specifically SFA which always takes more time than SAX. It is worth mentioning that the number of pairs generated using SFA is mostly greater than SAX. Both are generated using cross-validation on the training data. Therefore, the size of the training data is crucial for the hyper-parameterisation process. Chinatown is one of the smallest datasets in terms of length and size. Total time reported to run Co-eye is less than 1 minute. An extreme dataset is HandOutlines with a length of 2709 and 1000 training instances. Co-eye total time is 563 seconds which is reasonable given the length and size of the dataset. The longest time is reported on the crop dataset which is one of the shortest datasets, yet the training size is the largest (7200 instances). The training size consequently increases the time for parameter selection. FordA has a factor of both, long series and a large number of instances in training. Hence, the total time is as long as the crop dataset (with a shorter length, but double training size). Both training and prediction times are very small across all datasets. The longest prediction time, of 45 seconds, is recorded for crop dataset, with a test size of 168000, which is approximately 267 milliseconds per time series in this dataset. 

\begin{table}[!htp]
\caption{\color{black}Running time in seconds for each phase in Co-eye on selected datasets}
\begin{tabular}{p{1.5cm}p{0.5cm}p{0.5cm}p{0.5cm}p{1cm}p{1cm}p{1cm}p{1cm}p{1cm}}
\hline 
Dataset               & Train & Test  & Length & \multicolumn{2}{k}{Parameter Selection time} & Training time & Prediction time & Total time \\ \hline

                      &       &       &        & SAX                   & SFA                  &               &                 &            \\ \hline
Chinatown             & 20    & 345   & 24     & 13                 & 37              & 5        & 0.5          & 57      \\
ItalyPowDem&67&1029&45&12&49&7.2&0.9&71 \\

SonySurf1 & 20    & 601   & 70     & 13                & 174              & 13        & 1.5           & 204     \\
FordA                 & 3601  & 1320  & 500    & 185                & 968             & 132        & 5.2        & 1298  \\
HandOutlines          & 1000  & 370   & 2709   & 81             & 370               & 98         & 9.7           & 563     \\
Crop                  & 7200  & 16800 & 24     & 124                & 873& 168       & 45           & 1312 \\\hline
\end{tabular}
\label{tt}

\end{table}

\color{black}

\subsection{Case Study}
\label{casestudy}
\begin{figure}[!h]
    \centering
    \includegraphics[width=0.6\textwidth]{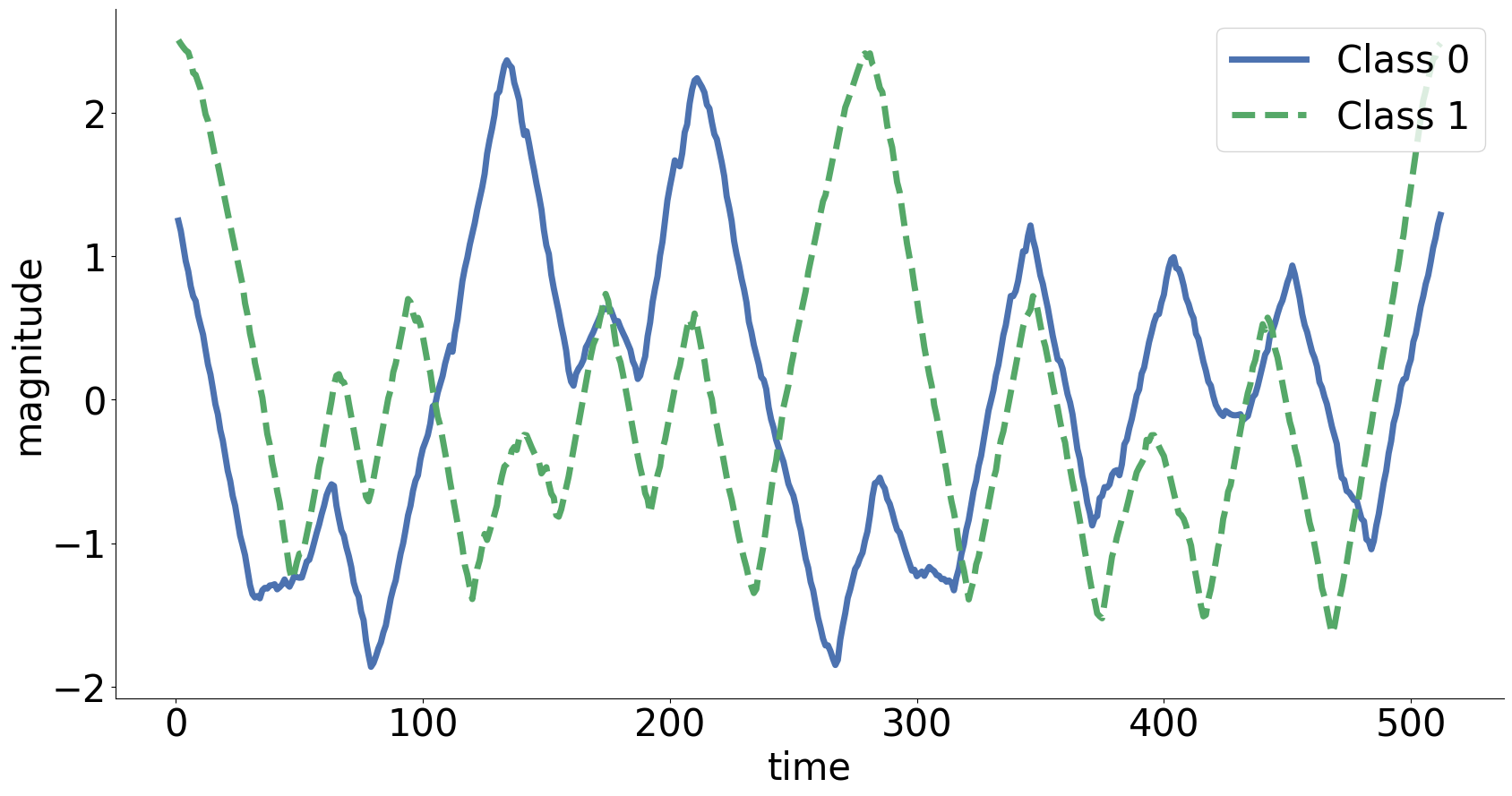}
    \caption {Two samples of the two classes in BeetleFly dataset}
    \label{CS1_TS}
    \end{figure}

The experimental work reveals a dataset that best matches the strengths of Co-eye. Accordingly, this dataset is used as a case study to illustrate how Co-eye performs, and its diverse granularity and dynamism make the technique of choice for some TSC problems. ``BettleFly'' dataset is used for testing contour/image and skeleton-based descriptors. Classes of images vary broadly, and include classes that are similar in shape to one another. There are 20 instances of each class, and 40 instances in total. Outlines of these images have been extracted and mapped into 1-D series of distances to the centre of length 512. Beetle/Fly is the problem of distinguishing between an outline of a beetle and a fly. Figure \ref{CS1_TS} shows two test samples representing each class; Beetle and Fly. 

\begin{figure}[!h]
\centering
\includegraphics[width=0.6\textwidth]{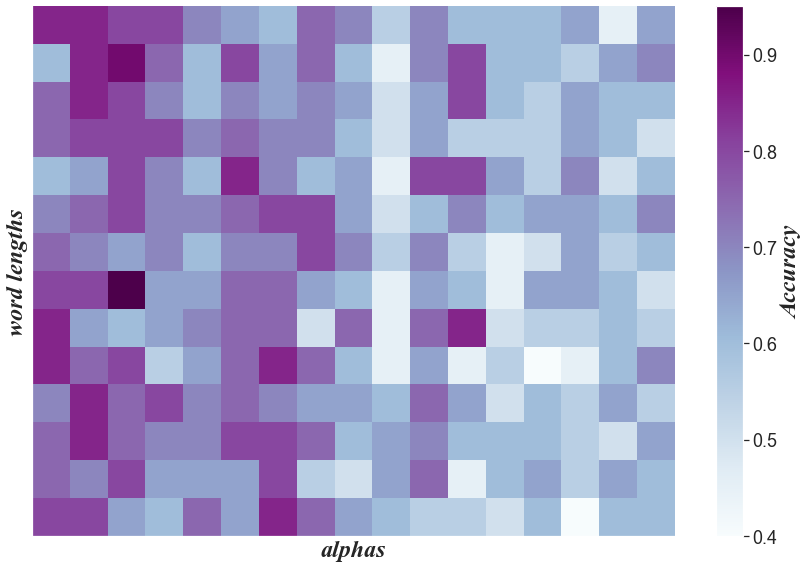}
\caption{Eye accuracy variations in BeetleFly dataset}
\label{HM}
\end{figure}

Co-eye has reported an accuracy of 100\% on this dataset which is the best among all methods in the literature so far. We first explore how each eye is performing individually, without combining them into a compound eye. Figure \ref{HM} shows the variation in accuracy for each pair of $\alpha$ and $w$. According to this matrix, no single eye reached the accuracy of a compound eye. The best accuracy reported 
is 95\% compared to 100\% with Co-eye. It is also noted that small $\alpha$ performs better for this dataset across all word lengths. This is consistent with the shape of the time series as shown in Figure \ref{CS1_TS}. 

\begin{figure}[!h]
\centering
\begin{subfigure}{\textwidth}
  \centering
  \includegraphics[scale=0.25]{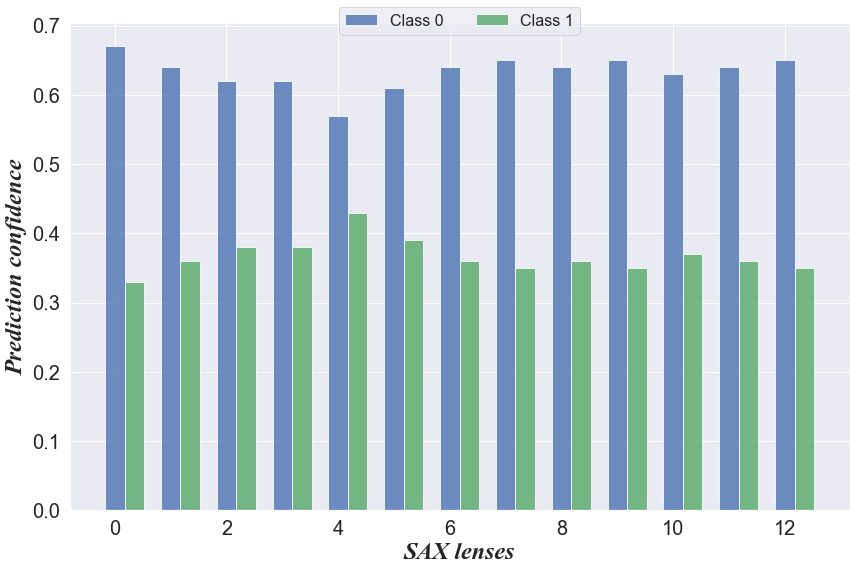}
  \caption{SAX lenses confidence for each class}
  \label{CS0SAX}
\end{subfigure}

\begin{subfigure}{\textwidth}
  \centering
  \includegraphics[scale=0.25]{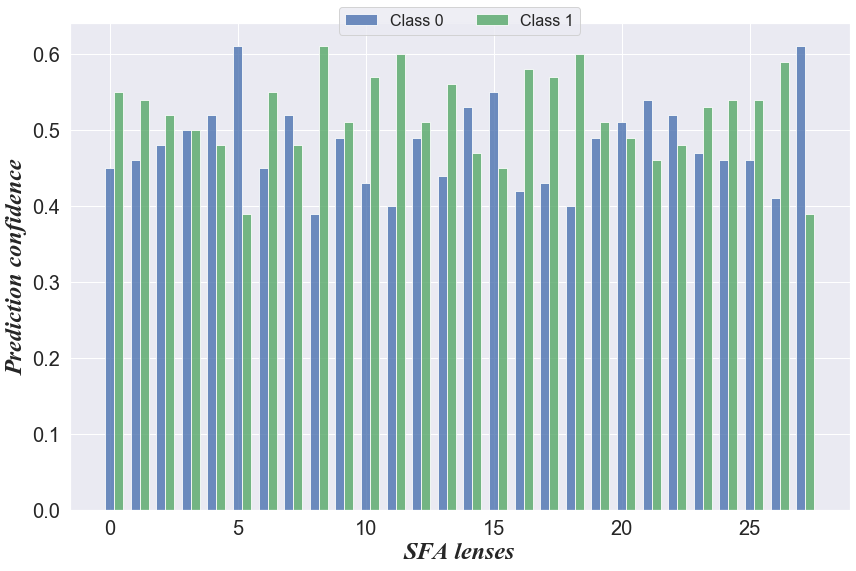}
  \caption{SFA lenses confidence for each class}
  \label{CS0SFA}
\end{subfigure}
    \caption{Prediction confidence of Co-eye lenses with True class 0}
    \label{class0_coeye}

\end{figure}

\textcolor{black}{Co-eye extracts a total of 41 lenses, 27 from SFA representation and 14 from SAX. Figure \ref{class0_coeye} displays the probability prediction for each lens with a single instance of class 0 (solid line in Figure \ref{CS1_TS}). The charts show the confidence associated to each class prediction using lenses from SAX representation in Figure \ref{CS0SAX} and SFA representation in Figure \ref{CS0SFA}. }

\begin{figure}[H]
\centering

\begin{subfigure}{\textwidth}
  \centering
  \includegraphics[scale=0.25]{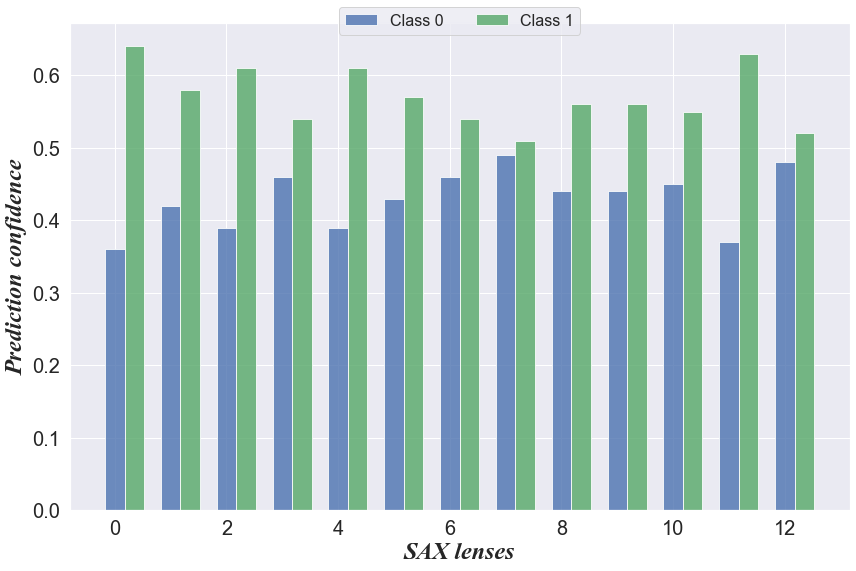}
  \caption{SAX lenses confidence for each class}
  \label{CS1SAX}
\end{subfigure}
\begin{subfigure}{\textwidth}
  \centering
  \includegraphics[scale=0.25]{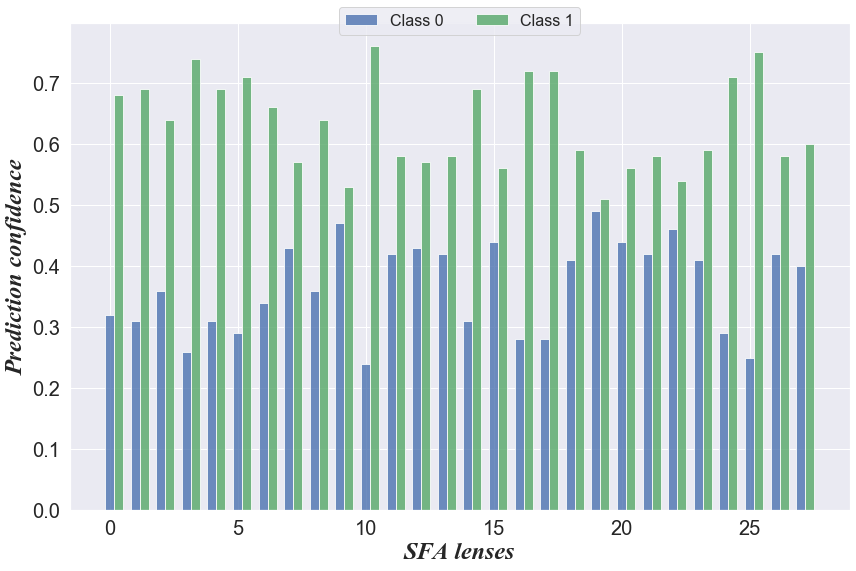}
  \caption{SFA lenses confidence for each class}
  \label{CS1SFA}
\end{subfigure}
    \caption{Prediction confidence of Co-eye lenses with True class 1}
    \label{class1_coeye}

\end{figure}

\textcolor{black}{As shown in the charts, SAX lenses can predict the correct class with high confidence. Among all SAX lenses/forests, the most confident lens/forest has $\alpha = 3$ and uniform segments (indexed 0 in Figure \ref{CS0SAX}). On the other hand, SFA lenses show uncertainty in classification between the two classes. The highest confidence of 0.61 is reported with the pairs ($w$= 20, $\alpha$= 7) and ($w$= 40, $\alpha$= 8) indexed 5 and 8, respectively, in Figure \ref{CS0SFA}. Although both  SFA lenses have the same confidence, they vote for different classes. With a tie in the SFA decision, while SAX's best lens votes for class 0, the prediction is settled to class 0 which is the true prediction. It is worth noting that alpha size in both representations suggests that a wider lens range between 3 and 8, but not very wide, is more significant to classify this class.
}
\textcolor{black}{Prediction confidence of the second class sample (dotted line in  Figure \ref{CS1_TS}) is depicted in Figure \ref{class1_coeye}. The charts show another disagreement between SAX and SFA predictions for the same dataset, but on a different class. SFA in general is more confident towards the correct prediction for this sample. This is opposite to the lack of confidence in SAX lenses/forests. The switch of importance between SAX and SFA confidence between Figure \ref{class0_coeye} and Figure \ref{class1_coeye} shows the importance of ensembling both representations to attain a broader view of the data in both frequency and time domains.}

``Win some, lose some'' is not the aim of these experiments rather than understanding when we win and when we lose. According to the aforementioned analysis, Co-eye demonstrates its best performance with datasets that have no significant phase shift such as spectrum, spectro, HRM, ECG and energy data. This is due to the approximate alignment of selected lenses and their corresponding forests. Whenever this alignment is significantly shifted, the lenses will be confused. An example of this confusion occurs when using a magnifying lens while looking for a global pattern that requires rather a wide lens. The results also show the significance of bringing together a multi-resolution view of the data across time and frequency domains, with combined accuracy better than each individual component.

\section {Conclusion and Future work}
\label{con}
 In this work, we have introduced Co-eye, our multi-resolution ensemble method for time series classification. Inspired by flies' compound eye, Co-eye brings together different lenses with multi-resolutions for broader visibility that covers both local and global views. Co-eye targets the diversification of time series by combining both time and frequency features using both SAX and SFA, respectively. In the evaluation, we conducted our experiments on the extended version of UCR repository with longer and more challenging datasets. The experiments show that Co-eye has a competitive accuracy compared to state-of-the-art techniques. Co-eye most significant accuracy improvement is attained in datasets that have no significant phase shifting such as spectrum and ECG. 
 
In future work, we explore in many directions. First, we investigate enhancing Co-eye performance with datasets that contain a significant phase shifting. This can be implemented by an initial alignment of series before applying Co-eye, however, that might increase the overall complexity. Second, Co-eye currently assumes fixed length of series, we aim in the future work to extend Co-eye to classify time series of varied lengths. We also aim to explore applying Co-eye to multidimensional time series. Finally, another notably successful tree-based ensemble method in TSC, namely, rotation forest, will be used as an alternative to random forests. 
\bibliographystyle{spmpsci}  
\bibliography{Coeye} 

\begin{thebibliography}{10}
\providecommand{\url}[1]{{#1}}
\providecommand{\urlprefix}{URL }
\expandafter\ifx\csname urlstyle\endcsname\relax
  \providecommand{\doi}[1]{DOI~\discretionary{}{}{}#1}\else
  \providecommand{\doi}{DOI~\discretionary{}{}{}\begingroup
  \urlstyle{rm}\Url}\fi

\bibitem{TSCwebsite}
Anthony~Bagnall Jason~Lines, W.V., Keogh, E.: The {UEA} and {UCR} time series
  classification repository.
\newblock \urlprefix\url{www.timeseriesclassification.com}

\bibitem{bagnall2017great}
Bagnall, A., Lines, J., Bostrom, A., Large, J., Keogh, E.: The great time
  series classification bake off: a review and experimental evaluation of
  recent algorithmic advances.
\newblock Data Mining and Knowledge Discovery \textbf{31}(3), 606--660 (2017)

\bibitem{bagnall2015time}
Bagnall, A., Lines, J., Hills, J., Bostrom, A.: Time-series classification with
  cote: the collective of transformation-based ensembles.
\newblock IEEE Transactions on Knowledge and Data Engineering \textbf{27}(9),
  2522--2535 (2015)

\bibitem{baydogan2013bag}
Baydogan, M.G., Runger, G., Tuv, E.: A bag-of-features framework to classify
  time series.
\newblock IEEE transactions on pattern analysis and machine intelligence
  \textbf{35}(11), 2796--2802 (2013)

\bibitem{bergstra2012random}
Bergstra, J., Bengio, Y.: Random search for hyper-parameter optimization.
\newblock Journal of Machine Learning Research \textbf{13}(Feb), 281--305
  (2012)

\bibitem{chawla2009data}
Chawla, N.V.: Data mining for imbalanced datasets: An overview.
\newblock In: Data mining and knowledge discovery handbook, pp. 875--886.
  Springer (2009)

\bibitem{chawla2002smote}
Chawla, N.V., Bowyer, K.W., Hall, L.O., Kegelmeyer, W.P.: Smote: synthetic
  minority over-sampling technique.
\newblock Journal of artificial intelligence research \textbf{16}, 321--357
  (2002)

\bibitem{UCR}
Dau, H.A., Keogh, E., Kamgar, K., Yeh, C.C.M., Zhu, Y., Gharghabi, S.,
  Ratanamahatana, C.A., Yanping, Hu, B., Begum, N., Bagnall, A., Mueen, A.,
  Batista, G.: The {UCR} time series classification archive (2018).
\newblock \url{https://www.cs.ucr.edu/~eamonn/time_series_data_2018/}

\bibitem{demvsar2006statistical}
Dem{\v{s}}ar, J.: Statistical comparisons of classifiers over multiple data
  sets.
\newblock Journal of Machine learning research \textbf{7}(Jan), 1--30 (2006)

\bibitem{deng2013time}
Deng, H., Runger, G., Tuv, E., Vladimir, M.: A time series forest for
  classification and feature extraction.
\newblock Information Sciences \textbf{239}, 142--153 (2013)

\bibitem{fawaz2019deep}
Fawaz, H.I., Forestier, G., Weber, J., Idoumghar, L., Muller, P.A.: Deep
  learning for time series classification: a review.
\newblock Data Mining and Knowledge Discovery \textbf{33}(4), 917--963 (2019)

\bibitem{finkel2003direct}
Finkel, D.E.: Direct optimization algorithm user guide.
\newblock Center for Research in Scientific Computation, North Carolina State
  University \textbf{2}, 1--14 (2003)

\bibitem{grabocka2014learning}
Grabocka, J., Schilling, N., Wistuba, M., Schmidt-Thieme, L.: Learning
  time-series shapelets.
\newblock In: Proceedings of the 20th ACM SIGKDD International Conference on
  Knowledge Discovery and Data Mining, KDD '14, pp. 392--401. ACM, New York,
  NY, USA (2014).
\newblock \doi{10.1145/2623330.2623613}

\bibitem{RandomForest}
Ho, T.K.: Random decision forests.
\newblock In: Proceedings of 3rd International Conference on Document Analysis
  and Recognition, vol.~1, pp. 278--282 vol.1 (1995).
\newblock \doi{10.1109/ICDAR.1995.598994}

\bibitem{Strawberry_dataset}
Holland, J.K., Kemsley, E.K., Wilson, R.H.: Use of fourier transform infrared
  spectroscopy and partial least squares regression for the detection of
  adulteration of strawberry purées.
\newblock Journal of the Science of Food and Agriculture \textbf{76}(2),
  263--269 (1998)

\bibitem{keogh2001dimensionality}
Keogh, E., Chakrabarti, K., Pazzani, M., Mehrotra, S.: Dimensionality reduction
  for fast similarity search in large time series databases.
\newblock Knowledge and information Systems \textbf{3}(3), 263--286 (2001)

\bibitem{keogh2000simple}
Keogh, E.J., Pazzani, M.J.: A simple dimensionality reduction technique for
  fast similarity search in large time series databases.
\newblock In: Proceedings of the 4th Pacific-Asia Conference on Knowledge
  Discovery and Data Mining, Current Issues and New Applications, PADKK '00,
  pp. 122--133. Springer-Verlag, London, UK (2000)

\bibitem{li2016multi}
Li, S., Li, Y., Fu, Y.: Multi-view time series classification: A discriminative
  bilinear projection approach.
\newblock In: Proceedings of the 25th ACM International on Conference on
  Information and Knowledge Management, pp. 989--998. ACM (2016)

\bibitem{lin2012rotation}
Lin, J., Khade, R., Li, Y.: Rotation-invariant similarity in time series using
  bag-of-patterns representation.
\newblock Journal of Intelligent Information Systems \textbf{39}(2), 287--315
  (2012)

\bibitem{lines2015time}
Lines, J., Bagnall, A.: Time series classification with ensembles of elastic
  distance measures.
\newblock Data Mining and Knowledge Discovery \textbf{29}(3), 565--592 (2015)

\bibitem{lines2016hive}
Lines, J., Taylor, S., Bagnall, A.: Hive-cote: The hierarchical vote collective
  of transformation-based ensembles for time series classification.
\newblock In: 2016 IEEE 16th International Conference on Data Mining (ICDM),
  pp. 1041--1046. IEEE (2016).
\newblock \doi{10.1109/ICDM.2016.0133}

\bibitem{patel2002mining}
Patel, P., Keogh, E., Lin, J., Lonardi, S.: Mining motifs in massive time
  series databases.
\newblock In: 2002 IEEE International Conference on Data Mining, 2002.
  Proceedings., pp. 370--377 (2002).
\newblock \doi{10.1109/ICDM.2002.1183925}

\bibitem{rodriguez2006rotation}
Rodriguez, J.J., Kuncheva, L.I., Alonso, C.J.: Rotation forest: A new
  classifier ensemble method.
\newblock IEEE transactions on pattern analysis and machine intelligence
  \textbf{28}(10), 1619--1630 (2006)

\bibitem{schafer2015boss}
Sch\"{a}fer, P.: The {BOSS} is concerned with time series classification in the
  presence of noise.
\newblock Data Mining and Knowledge Discovery \textbf{29}(6), 1505--1530
  (2015).
\newblock \doi{10.1007/s10618-014-0377-7}

\bibitem{schafer2012sfa}
Sch{\"a}fer, P., H{\"o}gqvist, M.: {SFA}: a symbolic {F}ourier approximation
  and index for similarity search in high dimensional datasets.
\newblock In: Proceedings of the 15th International Conference on Extending
  Database Technology, pp. 516--527. ACM (2012)

\bibitem{senin2013sax}
Senin, P., Malinchik, S.: Sax-vsm: Interpretable time series classification
  using sax and vector space model.
\newblock In: 2013 IEEE 13th International Conference on Data Mining, pp.
  1175--1180 (2013).
\newblock \doi{10.1109/ICDM.2013.52}

\bibitem{silva2013noninvasive}
Silva, I., Behar, J., Sameni, R., Zhu, T., Oster, J., Clifford, G.D., Moody,
  G.B.: Noninvasive fetal ecg: the physionet/computing in cardiology challenge
  2013.
\newblock In: Computing in Cardiology 2013, pp. 149--152. IEEE (2013)

\bibitem{wozniak2014survey}
Wo{\'z}niak, M., Gra{\~n}a, M., Corchado, E.: A survey of multiple classifier
  systems as hybrid systems.
\newblock Information Fusion \textbf{16}, 3--17 (2014)

\end{thebibliography}

\newpage
\appendix
\section{Classification accuracy on UCR repository}

   \footnotesize

\begin{longtable}{|p{1.8cm}p{0.5cm}p{0.5cm}p{0.5cm}p{0.1cm}p{0.5cm}|p{0.4cm}p{0.4cm}p{0.4cm}p{0.4cm}p{0.4cm}p{0.4cm}p{0.4cm}|}

\caption{Comparison of Co-eye accuracy with other state-of-the-art methods on UCR repository}
\label{allTable}

\\\hline
Dataset  & Type      & Train & Test & cls & Len & Co-eye & BOSS  & RandF & RotF & BoP  & SAX-VSM & LS   \\ \hline

\endfirsthead

\multicolumn{13}{c}%
{{\bfseries \tablename\ \thetable{} -- continued from previous page}} \\ \hline
Dataset                & Type      & Train & Test & cls & Len & Co-eye & BOSS   & RandF & RotF & BoP  & SAX-VSM & LS   \\ \hline
\endhead

\hline \multicolumn{13}{|r|}{{Continued on next page}} \\ \hline
\endfoot

\hline
\endlastfoot

Adiac                  & Image     & 390   & 391  & 37    & 176    & 0.77   & 0.76 & 0.64  & 0.77 & 0.59 & 0.45   & 0.52 \\
ArrowHead              & Image     & 36    & 175  & 3     & 251    & 0.8    & 0.83 & 0.71  & 0.74 & 0.74 & 0.79   & 0.85 \\
Beef                   & Spectro   & 30    & 30   & 5     & 470    & 0.83   & 0.8  & 0.73  & 0.87 & 0.6  & 0.43   & 0.87 \\
BeetleFly              & Image     & 20    & 20   & 2     & 512    & 1      & 0.9  & 0.8   & 0.9  & 0.7  & 0.9    & 0.8  \\
BirdChicken            & Image     & 20    & 20   & 2     & 512    & 0.6    & 0.95 & 0.75  & 0.85 & 0.75 & 1      & 0.8  \\
Car                    & Sensor    & 60    & 60   & 4     & 577    & 0.77   & 0.83 & 0.67  & 0.8  & 0.78 & 0.87   & 0.77 \\
CBF                    & Sim & 30    & 900  & 3     & 128    & 0.97   & 1    & 0.89  & 0.93 & 0.92 & 0.96   & 0.99 \\
ChlorineConc           & Sensor    & 467   & 3840 & 3     & 166    & 0.69   & 0.66 & 0.71  & 0.85 & 0.64 & 0.65   & 0.59 \\
CinCECGtorso           & Sensor    & 40    & 1380 & 4     & 1639   & 0.8    & 0.89 & 0.74  & 0.81 & 0.71 & 0.77   & 0.87 \\
Coffee                 & Spectro   & 28    & 28   & 2     & 286    & 1      & 1    & 1     & 1    & 0.93 & 0.93   & 1    \\
Computers              & Device    & 250   & 250  & 2     & 720    & 0.68   & 0.76 & 0.61  & 0.7  & 0.67 & 0.62   & 0.58 \\
CricketX               & Motion    & 390   & 390  & 12    & 300    & 0.59   & 0.74 & 0.59  & 0.63 & 0.58 & 0.74   & 0.74 \\
CricketY               & Motion    & 390   & 390  & 12    & 300    & 0.61   & 0.75 & 0.61  & 0.61 & 0.63 & 0.62   & 0.72 \\
CricketZ               & Motion    & 390   & 390  & 12    & 300    & 0.6    & 0.75 & 0.61  & 0.66 & 0.54 & 0.71   & 0.74 \\
DiaSizeRed    & Image     & 16    & 306  & 4     & 345    & 0.89   & 0.93 & 0.87  & 0.87 & 0.92 & 0.88   & 0.98 \\
DisPhalOutlCor         & Image     & 400   & 139  & 3     & 80     & 0.74   & 0.73 & 0.78  & 0.76 & 0.71 & 0.73   & 0.78 \\
DisPhalaOutlAgeG       & Image     & 600   & 276  & 2     & 80     & 0.68   & 0.75 & 0.77  & 0.75 & 0.69 & 0.84   & 0.72 \\
DistPhaTW              & Image     & 400   & 139  & 6     & 80     & 0.64   & 0.68 & 0.71  & 0.71 & 0.7  & 0.6    & 0.63 \\
Earthquakes            & Sensor    & 322   & 139  & 2     & 512    & 0.76   & 0.75 & 0.75  & 0.75 & 0.74 & 0.75   & 0.74 \\
ECG200                 & ECG       & 100   & 100  & 2     & 96     & 0.88   & 0.87 & 0.79  & 0.85 & 0.83 & 0.85   & 0.88 \\
ECG5000                & ECG       & 500   & 4500 & 5     & 140    & 0.94   & 0.94 & 0.93  & 0.95 & 0.91 & 0.91   & 0.93 \\
ECGFiveDays            & ECG       & 23    & 861  & 2     & 136    & 0.9    & 1    & 0.72  & 0.91 & 0.99 & 0.95   & 1    \\
ElectricDevices        & Device    & 8926  & 7711 & 7     & 96     & 0.69   & 0.8  & 0.65  & 0.79 & 0.6  & 0.71   & 0.59 \\
FaceAll                & Image     & 560   & 1690 & 14    & 131    & 0.84   & 0.78 & 0.73  & 0.91 & 0.76 & 0.97   & 0.75 \\
FaceFour               & Image     & 24    & 88   & 4     & 350    & 0.85   & 1    & 0.85  & 0.82 & 0.95 & 0.93   & 0.97 \\
FacesUCR               & Image     & 200   & 2050 & 14    & 131    & 0.82   & 0.96 & 0.78  & 0.8  & 0.9  & 0.93   & 0.94 \\
FiftyWords             & Image     & 450   & 455  & 50    & 270    & 0.69   & 0.71 & 0.68  & 0.66 & 0.54 & 0.39   & 0.73 \\
Fish                   & Image     & 175   & 175  & 7     & 463    & 0.84   & 0.99 & 0.76  & 0.83 & 0.9  & 0.95   & 0.96 \\
FordA                  & Sensor    & 3601  & 1320 & 2     & 500    & 0.92   & 0.93 & 0.77  & 0.84 & 0.79 & 0.83   & 0.96 \\
FordB                  & Sensor    & 3636  & 810  & 2     & 500    & 0.67   & 0.71 & 0.65  & 0.77 & 0.62 & 0.75   & 0.92 \\
GunPoint               & Motion    & 50    & 150  & 2     & 150    & 0.95   & 1    & 0.94  & 0.92 & 0.99 & 0.99   & 1    \\
Ham                    & Spectro   & 109   & 105  & 2     & 431    & 0.78   & 0.67 & 0.71  & 0.71 & 0.59 & 0.81   & 0.67 \\
HandOutlines           & Image     & 1000  & 370  & 2     & 2709   & 0.9    & 0.9  & 0.91  & 0.91 & 0.9  & 0.91   & 0.48 \\
Haptics                & Motion    & 155   & 308  & 5     & 1092   & 0.47   & 0.46 & 0.45  & 0.44 & 0.42 & 0.38   & 0.47 \\
Herring                & Image     & 64    & 64   & 2     & 512    & 0.58   & 0.55 & 0.59  & 0.66 & 0.56 & 0.63   & 0.63 \\
InlineSkate            & Motion    & 100   & 550  & 7     & 1882   & 0.3    & 0.52 & 0.35  & 0.37 & 0.38 & 0.41   & 0.44 \\
InsWingbtSound         & Sensor    & 220   & 1980 & 11    & 256    & 0.64   & 0.52 & 0.66  & 0.64 & 0.49 & 0.55   & 0.61 \\
ItalyPowDemand       & Sensor    & 67    & 1029 & 2     & 24     & 0.96   & 0.91 & 0.97  & 0.97 & 0.86 & 0.82   & 0.96 \\
lrgKitApp & Device    & 375   & 375  & 3     & 720    & 0.65   & 0.77 & 0.57  & 0.61 & 0.62 & 0.88   & 0.7  \\
Lightning2             & Sensor    & 60    & 61   & 2     & 637    & 0.79   & 0.84 & 0.74  & 0.69 & 0.75 & 0.85   & 0.82 \\
Lightning7             & Sensor    & 70    & 73   & 7     & 319    & 0.68   & 0.68 & 0.7   & 0.73 & 0.58 & 0.58   & 0.79 \\
Mallat                 & Sim & 55    & 2345 & 8     & 1024   & 0.96   & 0.94 & 0.86  & 0.95 & 0.85 & 0.84   & 0.95 \\
Meat                   & Spectro   & 60    & 60   & 3     & 448    & 0.93   & 0.9  & 0.93  & 0.97 & 0.73 & 0.9    & 0.73 \\
MedicalImages          & Image     & 381   & 760  & 10    & 99     & 0.66   & 0.72 & 0.72  & 0.77 & 0.49 & 0.51   & 0.66 \\
MidPhaOutlCor          & Image     & 400   & 154  & 3     & 80     & 0.74   & 0.78 & 0.81  & 0.8  & 0.71 & 0.68   & 0.78 \\
MidPhalOutlAgeG        & Image     & 600   & 291  & 2     & 80     & 0.55   & 0.55 & 0.6   & 0.57 & 0.52 & 0.55   & 0.57 \\
MiddlePhalanxTW        & Image     & 399   & 154  & 6     & 80     & 0.53   & 0.55 & 0.56  & 0.63 & 0.5  & 0.49   & 0.51 \\
MoteStrain             & Sensor    & 20    & 1252 & 2     & 84     & 0.88   & 0.88 & 0.89  & 0.88 & 0.85 & 0.79   & 0.88 \\
NinvFatECGTh1          & ECG       & 1800  & 1965 & 42    & 750    & 0.9    & 0.84 & 0.88  & 0.91 &      & 0.52   & 0.26 \\
NinvFatECGTh2         & ECG       & 1800  & 1965 & 42    & 750    & 0.92   & 0.9  & 0.91  & 0.92 &      & 0.59   & 0.77 \\
OliveOil               & Spectro   & 30    & 30   & 4     & 570    & 0.87   & 0.87 & 0.9   & 0.87 & 0.87 & 0.9    & 0.17 \\
OSULeaf                & Image     & 200   & 242  & 6     & 427    & 0.6    & 0.95 & 0.51  & 0.57 & 0.65 & 0.85   & 0.78 \\
PhalOutlCor            & Image     & 1800  & 858  & 2     & 80     & 0.76   & 0.77 & 0.82  & 0.86 & 0.71 & 0.71   & 0.76 \\
Phoneme                & Sensor    & 214   & 1896 & 39    & 1024   & 0.17   & 0.26 & 0.15  & 0.13 & 0.13 & 0.1    & 0.22 \\
Plane                  & Sensor    & 105   & 105  & 7     & 144    & 0.97   & 1    & 0.99  & 0.99 & 0.98 & 0.99   & 1    \\
ProxPhalOutCor         & Image     & 400   & 205  & 3     & 80     & 0.89   & 0.85 & 0.87  & 0.86 & 0.75 & 0.83   & 0.85 \\
ProxlPhaOutlAgeG       & Image     & 600   & 291  & 2     & 80     & 0.85   & 0.83 & 0.87  & 0.85 & 0.77 & 0.82   & 0.83 \\
ProxiPhalTW            & Image     & 400   & 205  & 6     & 80     & 0.73   & 0.8  & 0.81  & 0.82 & 0.66 & 0.61   & 0.78 \\
RefrigerationDev       & Device    & 375   & 375  & 3     & 720    & 0.53   & 0.5  & 0.51  & 0.57 & 0.46 & 0.65   & 0.51 \\
ScreenType             & Device    & 375   & 375  & 3     & 720    & 0.53   & 0.46 & 0.43  & 0.44 & 0.42 & 0.51   & 0.43 \\
ShapeletSim            & Sim& 20    & 180  & 2     & 500    & 0.63   & 1    & 0.49  & 0.41 & 0.7  & 0.72   & 0.95 \\
ShapesAll              & Image     & 600   & 600  & 60    & 512    & 0.76   & 0.91 & 0.77  & 0.74 & 0.79 & 0.7    & 0.77 \\
smlKitApp              & Device    & 375   & 375  & 3     & 720    & 0.67   & 0.73 & 0.67  & 0.73 & 0.75 & 0.58   & 0.66 \\
Sonysurf1  & Sensor    & 20    & 601  & 2     & 70     & 0.87   & 0.63 & 0.64  & 0.81 & 0.72 & 0.81   & 0.81 \\
Sonysurf2  & Sensor    & 27    & 953  & 2     & 65     & 0.84   & 0.86 & 0.79  & 0.81 & 0.81 & 0.82   & 0.88 \\
StarlightCurves        & Sensor    & 1000  & 8236 & 3     & 1024   & 0.96   & 0.98 & 0.95  & 0.97 &      & 0.88   & 0.95 \\
Strawberry             & Spectro   & 613   & 370  & 2     & 235    & 0.94   & 0.98 & 0.96  & 0.97 & 0.97 & 0.96   & 0.91 \\
SwedishLeaf            & Image     & 500   & 625  & 15    & 128    & 0.91   & 0.92 & 0.87  & 0.88 & 0.73 & 0.67   & 0.91 \\
Symbols                & Image     & 25    & 995  & 6     & 398    & 0.9    & 0.97 & 0.9   & 0.79 & 0.95 & 0.62   & 0.93 \\
SyntheticControl       & Sim & 300   & 300  & 6     & 60     & 0.98   & 0.97 & 0.94  & 0.97 & 0.93 & 0.89   & 1    \\
ToeSegmentation1       & Motion    & 40    & 228  & 2     & 277    & 0.71   & 0.94 & 0.65  & 0.53 & 0.88 & 0.93   & 0.93 \\
ToeSegmentation2       & Motion    & 36    & 130  & 2     & 343    & 0.73   & 0.96 & 0.77  & 0.58 & 0.95 & 0.86   & 0.92 \\
Trace                  & Sensor    & 100   & 100  & 4     & 275    & 0.98   & 1    & 0.78  & 0.93 & 0.97 & 1      & 1    \\
TwoLeadECG             & ECG       & 23    & 1139 & 2     & 82     & 0.7    & 0.98 & 0.72  & 0.97 & 0.98 & 0.9    & 1    \\
TwoPatterns            & Sim & 1000  & 4000 & 4     & 128    & 0.91   & 0.99 & 0.86  & 0.93 & 0.94 & 0.87   & 0.99 \\
UWaveGestLibX   & Motion    & 896   & 3582 & 8     & 945    & 0.76   & 0.76 & 0.76  & 0.78 & 0.6  & 0.53   & 0.79 \\
UWaveGestLibY   & Motion    & 896   & 3582 & 8     & 315    & 0.7    & 0.69 & 0.69  & 0.71 & 0.53 & 0.41   & 0.7  \\
UWaveGestLibZ   & Motion    & 896   & 3582 & 8     & 315    & 0.7    & 0.69 & 0.72  & 0.72 & 0.56 & 0.49   & 0.75 \\
UWaveGestLibAll & Motion    & 896   & 3582 & 8     & 315    & 0.95   & 0.94 & 0.94  & 0.94 &      & 0.8    & 0.95 \\
Wafer                  & Sensor    & 1000  & 6164 & 2     & 152    & 0.99   & 0.99 & 0.99  & 0.99 & 1    & 1      & 1    \\
Wine                   & Spectro   & 57    & 54   & 2     & 234    & 0.61   & 0.74 & 0.74  & 0.94 & 0.76 & 0.96   & 0.5  \\
WordSynonyms           & Image     & 267   & 638  & 25    & 270    & 0.58   & 0.64 & 0.57  & 0.6  & 0.52 & 0.49   & 0.61 \\
Worms                  & Motion    & 181   & 77   & 5     & 900    & 0.52   & 0.56 & 0.55  & 0.61 & 0.53 & 0.56   & 0.61 \\
WormsTwoClass          & Motion    & 181   & 77   & 2     & 900    & 0.71   & 0.83 & 0.64  & 0.69 & 0.64 & 0.71   & 0.73 \\
Yoga                   & Image     & 300   & 3000 & 2     & 426    & 0.82   & 0.92 & 0.81  & 0.82 & 0.83 & 0.84   & 0.83

\end{longtable}

\end{document}